%% file: main.tex
\pdfoutput=1

\documentclass[conference]{IEEEtran}

\usepackage{fancyhdr}
\usepackage[normalem]{ulem}
\usepackage[hyphens]{url}

\usepackage{algorithm}
\usepackage{algorithmic}
\usepackage{xspace}
\usepackage{amsmath}
\usepackage{mathptmx}
\usepackage{mathtools}
\usepackage{xcolor}
\usepackage{amsfonts}
\usepackage{amssymb}
\usepackage{pifont}
\usepackage{listings}             
\usepackage[caption=false]{subfig}

\newcommand{\ignore}[1]{}

\newcommand{\dflowdense}[0]{PT-IS-CP-dense\xspace}
\newcommand{\dflowsparse}[0]{PT-IS-CP-sparse\xspace}
\newcommand{\dflowsparselong}[0]{{\em PlanarTiled-InputStationary-CartesianProduct-sparse}\xspace}
\newcommand{\dflowdcnn}[0]{PT-IS-DP-dense\xspace}
\newcommand{\scnn}[0]{\texttt{SCNN}\xspace}

\newcommand{\orac}[0]{\texttt{SCNN(oracle)}\xspace}
\newcommand{\fig}[1]{Figure~\ref{#1}}
\newcommand{\sect}[1]{Section~\ref{#1}}

\newcommand{\avgPerf}[0]{2.7$\times$\xspace}                   
\newcommand{\avgEnergyScnn}[0]{2.3$\times$\xspace}             
\newcommand{\avgEnergyDcnnOpt}[0]{2.0$\times$\xspace}          

\newcommand{\perfAlexnet}[0]{2.37$\times$\xspace}              
\newcommand{\perfGooglenet}[0]{2.19$\times$\xspace}            
\newcommand{\perfVggnet}[0]{3.52$\times$\xspace}               

\newcommand{\colwidth}{0.45\textwidth}

\newcommand\blfootnote[1]{%
\begingroup
\renewcommand\thefootnote{}\footnote{#1}%
\addtocounter{footnote}{-1}%
\endgroup
}

\title{SCNN: An Accelerator for Compressed-sparse Convolutional Neural Networks}

\author{Angshuman Parashar$^\dagger$ \hspace{0.1in} Minsoo Rhu$^\dagger$ \vspace*{0.05in} \\ 
Anurag Mukkara$^\ddagger$ \hspace{0.1in} Antonio Puglielli$^\ast$ \hspace{0.1in} Rangharajan Venkatesan$^\dagger$ \hspace{0.1in} Brucek Khailany$^\dagger$ \hspace{0.1in} \vspace*{0.05in} \\  
Joel Emer$^{\dagger \ddagger}$ \hspace{0.1in} Stephen W. Keckler$^\dagger$ \hspace{0.1in} William J. Dally$^{\dagger \diamond}$ 
\vspace*{0.1in}
\\
NVIDIA$^\dagger$ \hspace*{0.1in} Massachusetts Institute of Technology$^\ddagger$ \hspace*{0.1in} UC-Berkeley$^\ast$ \hspace*{0.1in} Stanford University$^\diamond$ \vspace*{0.2in}
}

\begin{document}
\maketitle
\pagestyle{plain}


\input{tex/abs.tex}

\IEEEpeerreviewmaketitle

\blfootnote{
\hspace*{-0.12in}A version appears in the $44^{\text{th}}$ IEEE/ACM International Symposium on Computer Architecture (ISCA-44), 2017.
\\ \\
This research was, in part, funded by the U.S. Government, under the
DARPA CRAFT program. The views and conclusions contained in this
document are those of the authors and should not be interpreted as
representing the official policies, either expressed or implied, of
the U.S. Government.
}

\input{tex/intro.tex}

\input{tex/motivation.tex}

\input{tex/dataflow.tex}

\input{tex/arch.tex}

\input{tex/infrastructure.tex}

\input{tex/eval.tex}

\input{tex/related.tex}

\input{tex/conclusion.tex}


\bibliographystyle{abbrv}
\bibliography{scnn}

\end{document}

%% file: tex/abs.tex
\begin{abstract}

Convolutional Neural Networks (CNNs) have emerged as a fundamental
technology for machine learning.  High performance and extreme energy
efficiency are critical for deployments of CNNs in a wide range of
situations, especially mobile platforms such as autonomous vehicles,
cameras, and electronic personal assistants.  This paper introduces
the Sparse CNN (SCNN) accelerator architecture, which improves
performance and energy efficiency by exploiting the zero-valued
weights that stem from network pruning during training and zero-valued
activations that arise from the common ReLU operator applied during
inference.  Specifically, SCNN employs a novel dataflow that enables
maintaining the sparse weights and activations in a compressed
encoding, which eliminates unnecessary data transfers and reduces
storage requirements.  Furthermore, the SCNN dataflow facilitates
efficient delivery of those weights and activations to the multiplier
array, where they are extensively reused.  In addition, the
accumulation of multiplication products are performed in a novel
accumulator array.  Our results show that on contemporary neural
networks, SCNN can improve both performance and energy by a factor of
\avgPerf and \avgEnergyScnn, respectively, over a comparably
provisioned dense CNN accelerator.

\end{abstract}

%% file: tex/intro.tex
\section{Introduction}


Driven by the availability of massive data and the computational
capability to process it, deep learning has recently emerged as a
critical tool for solving complex problems across a wide range of
domains, including image recognition~\cite{alexnet}, speech
processing~\cite{graves:2005:fpc,deepspeech_1,deepspeech_2}, natural
language processing~\cite{collobert:2011:nlp_from_scratch}, language
translation~\cite{persistent_rnn}, and autonomous
vehicles~\cite{LeCun:Nature:2015}. Convolutional neural networks
(CNNs) have become the most popular algorithmic approach for deep
learning for many of these domains. Employing CNNs can be decomposed
into two tasks: (1) training \textemdash~in which the parameters of a
neural network are learned by observing massive numbers of training
examples, and (2) inference \textemdash~in which a trained neural
network is deployed in the field and classifies the observed data.
Today, training is often done on GPUs~\cite{cudnn} or farms of GPUs,
while inference depends on the application and can employ CPUs, GPUs,
FPGA, or specially-built ASICs\@.

During the training process, a deep learning expert will typically architect
the network, establishing the number of layers, the operation performed by each
layer, and the connectivity between layers.  Many layers have parameters,
typically filter weights, which determine their exact computation.  The
objective of the training process is to learn these weights, usually via a
stochastic gradient descent-based excursion through the space of weights. This
process typically employs a forward-propagation calculation for each training
example, a measurement of the error between the computed and desired output,
and then back-propagation through the network to update the weights.  Inference
has similarities, but only includes the forward-propagation calculation.
Nonetheless, the computation requirements for inference can be prohibitively
large, particularly with the emergence of deeper networks (hundreds of
layers~\cite{msr_2015,nn_stochastic_depth,rhu:2016:vdnn,wired:msr_deeper_nn}) and larger
inputs sets, such as high-definition video.  Furthermore, the energy efficiency
of this computation is important, especially for mobile platforms, such as
autonomous vehicles, cameras, and electronic personal assistants.


Recent published works have shown that common networks have significant
redundancy and can be pruned dramatically during training without substantively
affecting accuracy~\cite{song:2015:pruning}. Our experience shows that the
number of weights that can be eliminated varies widely across the layers but
typically ranges from 20\% to
80\%~\cite{song:2015:pruning,song:2015:compression}. Eliminating weights
results in a network with a substantial number of zero values, which can
potentially reduce the computational requirements of inference.
  
The inference computation also offers a further optimization opportunity. 
In specific, many networks employ as their non-linear operator the
ReLU (rectified linear unit) function which clamps all negative
activation values to zero. The activations are the output values of an
individual layer that are passed as inputs to the next layer. Our
experience shows that for typical data sets, 50--70\% of the
activations are clamped to zero. 
Since the multiplication of weights and activations is the key computation for inference,
the combination of these two factors can reduce the amount of computation required by over
an order of magnitude. 
Additional benefits can be achieved by a compressed encoding for zero weights and activations,
thus allowing more to fit in on-chip RAM and eliminating energy-costly DRAM accesses.


In this paper, we introduce the Sparse CNN (SCNN) accelerator
architecture, a new CNN inference architecture that exploits both
weight and activation sparsity to improve both performance and power.
Previous works have employed techniques for exploiting sparsity,
including saving computation energy for zero-valued activations
and compressing weights and activations stored in DRAM~\cite{EyerissISSCC:2016,EyerissISCA:2016}.
Other works have used a compressed encoding of activations~\cite{CnvlutinISCA:2016}
or weights~\cite{CambriconX:2016}
in parts of their dataflow to reduce data transfer bandwidth 
and save time for computations of some
multiplications with a zero operand.
While these prior architectures have largely focused on 
eliminating computations and exploiting some data compression,
SCNN couples an algorithmic dataflow that eliminates all multiplications with a zero operand
while employing a compressed representation of both weights and activations
through almost the entire computation.

At the heart of the SCNN design is a processing element (PE)
with a multiplier array that accepts a vector of weights and a vector of activations.
Unlike previous convolutional dataflows~\cite{DianNaoASPLOS:2014,ShiDianNaoISCA:2015,EyerissISSCC:2016,MinervaISCA:2016},
the SCNN dataflow only delivers weights and activations to the multiplier array
that can all be multiplied by one another 
in the manner of a {\em Cartesian product}.
Furthermore, the activation vectors are reused in an {\em input stationary}~\cite{EyerissISCA:2016} fashion
against a number of weight vectors to reduce data accesses.
Finally, only non-zero weights and activations are fetched from the input storage arrays 
and delivered to the multiplier array.
As with any CNN accelerator, SCNN must accumulate the partial products generated by the multipliers.
However, since the products generated by the multiplier array cannot be directly summed together,
SCNN tracks the output coordinates associated with each multiplication 
and sends the coordinate and product to a scatter accumulator array for summing. 

To increase parallelism beyond a single PE, multiple PEs can be run in parallel
with each working on a disjoint 3D {\em tile} of input activations.
Because of the end-to-end compression of activations,
SCNN can keep the both the input and output activations of each tile local to its PE, 
further reducing energy-hungry data transmission.
Overall, this results in a design with
efficient compressed storage and delivery of input operands,
high reuse of the input operands in the multiplier array,
and that spends no time on multiplications with zero operands.

To evaluate SCNN, we developed a cycle-level performance model and a
validated analytical model that allows us to quickly explore the
design space of different types of accelerators. We also implemented
an SCNN PE in synthesizable System C and compiled the design into
gates using a combination of commercial high-level synthesis (HLS)
tools and a traditional verilog compiler. Our results show that 64 PE
SCNN implementation with 16 multipliers per PE (1024 multipliers in
total) can be implemented in approximately 7.9$mm^2$, which is somewhat
larger than an equivalently provisioned dense accelerator architecture
due to the overheads of managing the sparse dataflow. On a range of
networks, SCNN provides a factor of \avgPerf speedup and a
\avgEnergyScnn energy reduction relative to the dense
architecture.

%
%


%% file: tex/motivation.tex
\section{Motivation}
\label{sec:motivation}

Convolutional Neural Network algorithms (CNNs) are essentially a
cascaded set of pattern recognition filters trained with
supervision~\cite{LeCun:Nature:2015}. A CNN consists of a series of
layers, which include convolutional layers, non-linear scalar operator
layers, and layers that downsample the intermediate data, for example
by pooling.  The convolutional layers represent the core of the CNN
computation and are characterized by a set of filters that are usually
$1{\times}1$ or $3{\times}3$, and occasionally $5{\times}5$ or larger.
The values of these filters are the {\em weights} that are trained
using a training set for the network.  Some deep neural networks
(DNNs) also include fully-connected layers, typically toward the end
of the DNN.  During inference, a new image (in the case of image
recognition) is presented to the network, which classifies into the
training categories by computing in succession each of the layers in
the network. The intermediate data between the layers are called {\em
  activations} and the output activation of one layer becomes the
input activation of the next layer. In this paper, we focus on
accelerating the convolutional layers as they constitute the majority
of the computation~\cite{Cong:CNNComputation:2014}.


Table~\ref{tab:networks} lists the attributes of three commonly used
networks in image processing: AlexNet~\cite{alexnet},
GoogLeNet~\cite{googlenet}, and VGGNet~\cite{vggnet}, whose
specifications come from the Caffe BVLC Model Zoo~\cite{CaffeZoo}. The
increasing layer depth across the networks represents the successively
more accurate networks in the ImageNet~\cite{imagenet} competition. The Maximum
Weights and Activations columns indicate the size of the
largest weight and activation matrices across the layer of the
network. The last column lists the total number of multiplies required
to compute a single inference pass through all of the convolutional
layers of the network. These data and computational requirements are
derived from the standard ImageNet inputs images of $224\times224$
pixels. Processing larger, higher resolution images will result in
greater computational and data requirements.


\begin{table}[tpb]
  \begin{center}
  \caption{Network characteristics. Weights and activations assume a data-type size of two bytes.}
\vspace*{0.1in}
  \begin{footnotesize}
  \renewcommand\tabcolsep{3pt}
  \begin{tabular}{|c|c|c|c|c|}
    \hline
    & \# Conv. & Max. Layer & Max. Layer & Total \# \\
    {\bf Network} & Layers & Weights & Activations & Multiplies \\
    \hline
    \hline
    AlexNet~\cite{alexnet}     & $5$  & $1.73$ MB  & $0.31$ MB  & $0.69$ B\\
    \hline                              
    GoogLeNet~\cite{googlenet} & $54$ & $1.32$ MB  & $1.52$ MB  & $1.1$ B\\
    \hline                              
    VGGNet~\cite{vggnet}       & $13$ & $4.49$ MB & $6.12$ MB & $15.3$ B\\
    \hline
  \end{tabular}
  \end{footnotesize}
  \end{center}
  \label{tab:networks}
\end{table}

{\bf Sparsity in CNNs.}
Sparsity in a layer of a CNN is defined as the fraction of zeros in
the layer's weight and input activation matrices. The primary technique
for creating weight sparsity is to prune the network during training.
Han, et al. developed a pruning algorithm that operates in two
phases~\cite{song:2015:pruning}. First, any weight with an absolute
value that is close to zero (e.g. below a defined threshold) is set to
zero. This process has the effect of removing weights from the filters,
and sometimes even forcing an output activation to always to be zero.
Second, the
remaining network is retrained, to regain the accuracy lost through
na\"{\i}ve pruning. The result is a smaller network with accuracy extremely
close to the original network. The process can be iteratively repeated
to reduce network size while maintaining accuracy.

Activation sparsity occurs dynamically during inference and is highly
dependent on the data being processed. Specifically, the rectified
linear unit (ReLU) function that is commonly used as the non-linear
operator in CNNs forces all negatively valued activations to be
clamped to zero. After completing computation of a convolutional
layer, a ReLU function is applied point-wise to each element in the
output activation matrices before the data is passed to the next
layer.



\begin{figure}[tbp] \centering
\subfloat[AlexNet]{
\includegraphics[width=\colwidth]{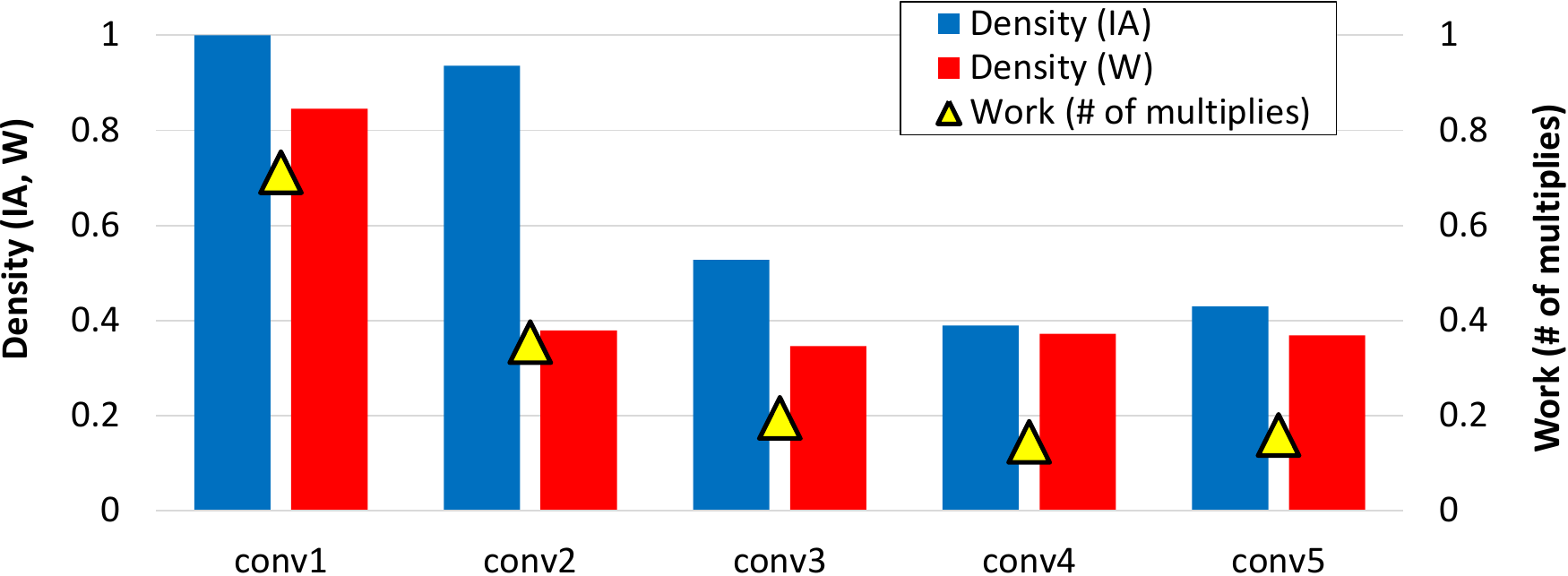}
\label{fig:density_sol_speedup_alexnet}
}
\vspace{1em}
\subfloat[GoogLeNet]{
\includegraphics[width=\colwidth]{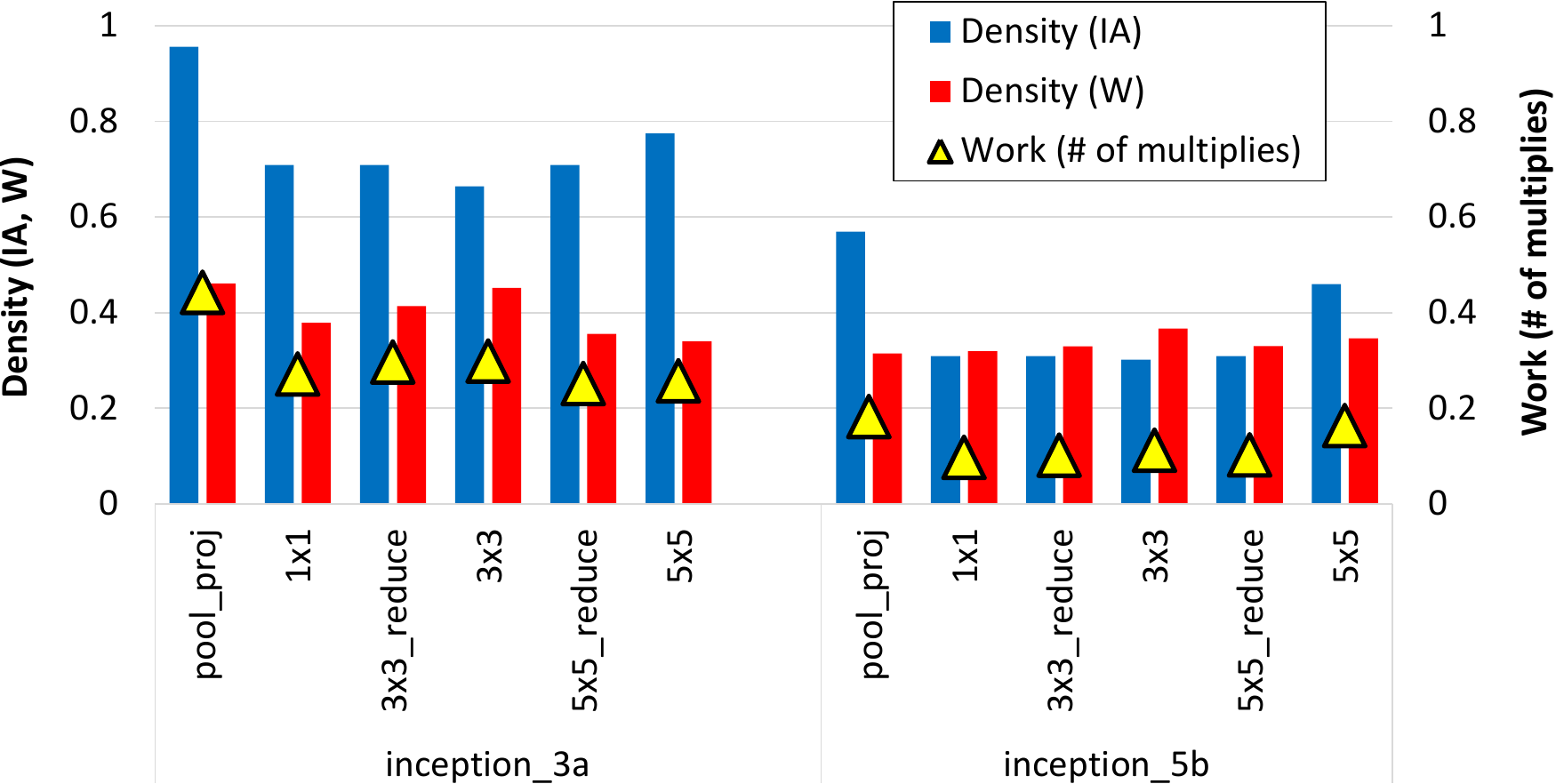}
\label{fig:density_sol_speedup_googlenet}
}
\vspace{1em}
\subfloat[VGGNet]{
\includegraphics[width=\colwidth]{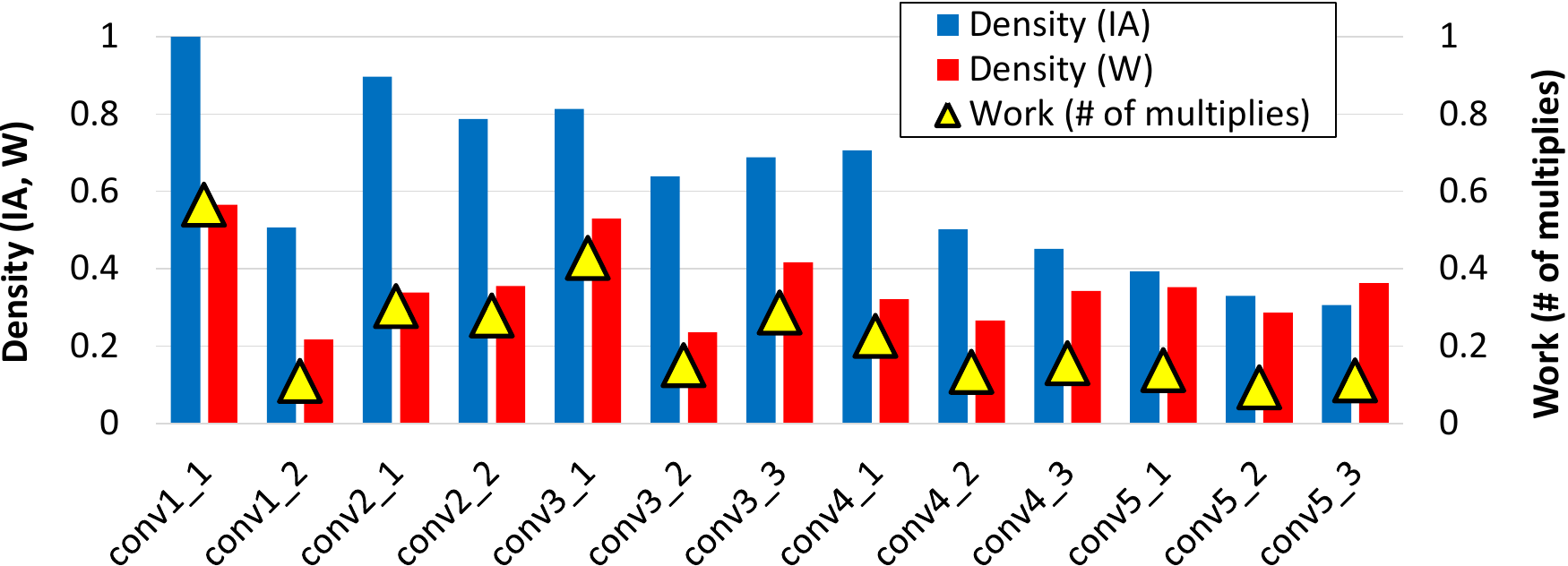}
\label{fig:density_sol_speedup_vgg}
}
\caption{Input activation and weight density and the reduction in the amount of work achievable by
maximally exploiting sparsity.}
\vspace{0em}
\label{fig:density_sol_speedup} 
\end{figure}

To measure the weight and activation sparsity, we used the Caffe
framework~\cite{caffe} to prune and train the three networks listed in
Table~\ref{tab:networks}, using the pruning algorithm
of~\cite{song:2015:pruning}. We then instrumented the Caffe framework
to inspect the activations between the convolutional
layers. Figure~\ref{fig:density_sol_speedup} shows the weight and
activation density (fraction of non-zeros or complement of sparsity)
of the layers of the networks, referenced to the left-hand y-axes. As
GoogLeNet has $54$ convolutional layers, we only show a subset of
representative layers. The data shows that weight density varies
across both layers and networks, reaching a minimum of 30\% for some
of the GoogLeNet layers. Activation density also varies, with density
typically being higher in early layers. Activation density can be as
low as 30\% as well. The triangles shows the ideal amount of work
(measured in multiplies of non-zero values) that could be achieved
through maximum exploitation of sparsity by taking the product of the
weight and activation densities on a per-layer basis. Typical layers
can reduce work by a factor of 4, and can reach as high as a factor of
ten.

{\bf Exploiting sparsity.}
Since multiplication by zero just results in a zero, it should require
no work.  In addition, that zero will contribute nothing to the
partial sum it is part of, so the addition is unnecessary as well.
Furthermore, data with many zeros can be represented in a compressed
form.  Together these characteristics provide a number of
opportunities for optimization:

\begin{itemize}

\item {\bf Compressing data:} Encoding the sparse weights and/or
  activations provides an architecture an opportunity to reduce the
  amount of data that must be moved throughout the memory hierarchy.
  It also reduces the data footprint, which allows larger matrices to
  be held a given size storage structure.

\item {\bf Eliminating computation:} For multiplications that have a
  zero weight and/or activation operand, the operation can be data
  gated or the operands might never be sent to the multiplier.  This
  can save energy consumption or both time and energy consumption,
  respectively.

\end{itemize}

Our SCNN architecture exploits both these opportunities.  First, it
employs a dense encoding of sparse weights and activations.  Second,
it uses a novel dataflow that delivers only those densely encoded
weights and activations to the multipliers.

%



%% file: tex/dataflow.tex
\section{SCNN Dataflow}
\label{sec:dataflow}

\begin{figure}[tbp]
\includegraphics[trim={0mm 2mm 3mm 0mm},clip,width=\columnwidth]{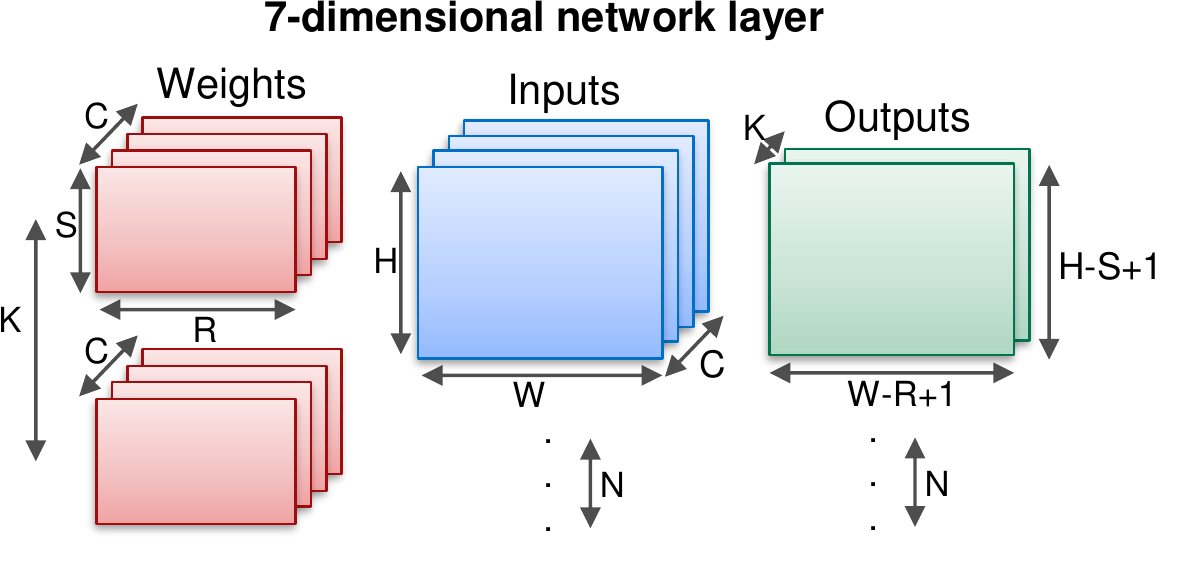}
\caption{CNN computations and parameters.}
\label{fig:CNN}
\end{figure}

The core operation in a CNN convolutional layer is a 2-dimensional sliding-window
convolution of an $R\times S$ element {\em filter} over a $W\times H$ element {\em
input activation} plane to produce a $W\times H$ element {\em output activation} plane. 
There can be multiple ($C$) input activation planes, which are referred to as {\em input channels}.
A distinct filter is applied to each input activation channel,
and the filter output for each of the $C$ channels are accumulated together element-wise into
a single output activation plane. Multiple filters ($K$) can be applied
to the same body of input activations to produce $K$ {\em output channels} of
activations. Finally, a batch of length $N$ of groups of $C$ channels of input
activation planes can be applied to the same volume of filter
weights. Figure~\ref{fig:CNN} shows these parameters applied to the
computation of a single CNN layer.

The set of computations for the complete layer
can be formulated as a loop nest over these 7 variables.
Because multiply-add operations are associative (modulo rounding errors,
which we will ignore in this study),
all permutations of these 7 loop variables are legal. 
Figure~\ref{fig:7dnest} shows an example loop nest based on one such permutation. 
We can concisely describe this nest as 
$N\rightarrow K\rightarrow C\rightarrow W\rightarrow H\rightarrow R\rightarrow S$.
Each point in the 7-dimensional space formed from these variables
represents a single multiply-accumulate operation.
Note that for the remainder of this paper,
we assume a batch size of 1, which is common for inferencing tasks.

This simple loop nest can be transformed in numerous ways to capture
different reuse patterns of the activations and weights and to map the
computation to a hardware accelerator implementation. A CNN's {\em
  dataflow} defines how the loops are ordered, partitioned, and
parallelized~\cite{EyerissISCA:2016}.  Prior
work\cite{EyerissISCA:2016} has shown that the choice of dataflow has
a significant impact on the area and energy-efficiency of an
architecture.  In fact, the choice of dataflow is perhaps the single
most significant differentiator between many prior works on CNN
architectures.

While the concept of dataflow has been studied for dense
architectures, sparse architectures can also employ various
alternative dataflows, each with its own set of trade-offs.  While an
exhaustive enumeration of sparse dataflows is beyond the scope of this
paper, we present a specific dataflow called \dflowsparselong
(or \dflowsparse). After examining a range of
different dataflows, we selected \dflowsparse because it enables reuse
patterns that exploit the characteristics of sparse weights and
activations. This section first presents an equivalent dense dataflow
(\dflowdense) to explain the decomposition of the computations and then
adds the specific features for \dflowsparse\@.

\subsection{The \dflowdense Dataflow}

{\bf Single-multiplier temporal dataflow.}  To understand the {\em
temporal} component of the \dflowdense dataflow, consider the
operation of a {\em processing element} (or PE) with a single
multiply-accumulate unit. The dataflow employs an {\em input-stationary}
computation order in which an input activation is held stationary at
the computation units as it is multiplied by 
all the filter weights needed to make all its contributions to each of the
the $K$ output channels (a $K\times R\times S$ subvolume). 
Thus each input activation will contribute to a volume of $K\times R\times S$ output activations.
This order maximizes the reuse of the input
activations, while paying a cost to stream the weights to the
computation units.  Accommodating multiple input channels ($C$) adds an
additional outer loop and results in the loop nest $C\rightarrow
W\rightarrow H\rightarrow K\rightarrow R\rightarrow S$.

\begin{figure}[tbp]
\begin{verbatim}
 for n = 1 to N
   for k = 1 to K
     for c = 1 to C
       for w = 1 to W
         for h = 1 to H
           for r = 1 to R
             for s = 1 to S
               out[n][k][w][h] +=
                 in[n][c][w+r-1][h+s-1] *
                 filter[k][c][r][s];
\end{verbatim}
\caption{7-dimensional CNN loop nest.}
\label{fig:7dnest}
\end{figure}

%
%
%

The \dflowdense dataflow requires input buffers for weights and input activations,
and a accumulator buffer to store the {\em partial sums} of the output activations.
The accumulator buffer must
perform a read-add-write operation for every access to a
previously-written index.  We call this accumulator buffer along with the
attached adder an {\em accumulation unit}.

Parameters of contemporary networks cause these buffers to be large
and energy-expensive to access.  The input-stationary temporal loop
nest amortizes the energy cost of accessing the input buffer over
multiple weight and accumulator buffer accesses.  More precisely, the
register in which the stationary input is held over $K\times R\times
S$ iterations serves as an inner buffer filtering accesses to the
larger input buffer.

Unfortunately, the stationarity of input activations comes at the cost
of more streaming accesses to the weights and partial sums (in the accumulator buffer).
Blocking the weights and partial sums in the output channel ($K$) dimension can
increase reuse of these data structures and improve energy efficiency.
We factor the output channel variable ($K$) into $K_c$ (which we call a {\em output-channel group})
and $K/K_c$, and only store weights and outputs for a single output-channel group at
a time inside the weight and accumulator buffers. Thus the subvolumes that
are housed in buffers at the computation unit are:

\begin{itemize}
\item Weights: $C\times K_c\times R\times S$
\item Inputs: $C\times W\times H$
\item Partial Sums: $K_c\times W\times H$
\end{itemize}

An outer loop over all the $K/K_c$ output-channel chunks results in
the complete loop nest $K/K_c\rightarrow C\rightarrow W\rightarrow
H\rightarrow K_c\rightarrow R\rightarrow S$.  Note that each iteration
of this outer loop will require the weight buffer to be refilled and the
accumulator buffer must be drained and cleared, while the contents of the input buffer will
be fully reused because the same input activations are used across all
output channels.


{\bf Intra-PE parallelism.}  To exploit the parallelism of many
multipliers within a PE, we fetch a vector of $F$ filter-weights from the
weight buffer and a vector of $I$ inputs from the input
activation buffer. These values are delivered to an array of
$F{\times}I$ multipliers to compute a full Cartesian product of output
partial-sums. Each product yields a useful partial sum such that no
extraneous fetches or computations are performed. \dflowsparse will
exploit this same property to make computation efficient on
compressed-sparse weights and input activations.
  
%
%



The multiplier outputs are sent to the accumulation unit, which
updates the partial sums of the output activation.  Each multiplier
output is accumulated with a partial sum at the matching output
coordinates in the output activation space. These coordinates are
computed in parallel with the multiplications.  The accumulation unit
must employ at least $F{\times}I$ adders to match the throughput of
the multipliers. Figure~\ref{fig:ptis-dense} shows pseudo-code for the
\dflowdense dataflow, including blocking in the $K$ dimension (A,C),
fetching vectors of input activations and weights (B,D), and computing
the Cartesian product in parallel (E,F)\@.  The $Kcoord()$,
$Xcoord()$, and $Ycoord()$ functions compute the $k$, $x$, and $y$
coordinates of the uncompressed output volume using a de-linearization
of the temporal loop indices $a$ and $w$, the spatial loop indices $i$
and $f$, and the known filter width and height.  Note that this
dataflow is simply a reordered, partitioned and parallelized version
of Figure~\ref{fig:7dnest}.


{\bf Inter-PE parallelism.} To scale beyond the practical limits of
multiplier count and buffer sizes within a PE, we employ a tiling
strategy to spread the work across an array of PEs so that each PE can
operate independently.  \dflowdense partitions the $W{\times}H$ element
activation plane into smaller $W_t{\times}H_t$ element {\em tiles} that are
distributed across the PEs\@. Each tile extends fully into the
input-channel dimension $C$, resulting in an input-activation volume
of $C{\times}W_t{\times}H_t$ assigned to each PE\@. Weights are
broadcast to the PEs and each PE operates on its own subset of the
input and output activation space. 

Unfortunately, strictly partitioning both input and output activations into
$W_t\times H_t$ tiles does not work because the sliding-window nature of
the convolution operation introduces cross-tile dependencies at tile edges.
These dependencies are called {\em halos}. Halos can be resolved in two
ways:
 
\begin{itemize}

\item {\bf Input halos:} The input buffers at each PE are sized to be slightly
  larger than $C\times W_t\times H_t$ to accommodate the halos. These halo
  input values are replicated across adjacent PEs, but outputs are strictly
  private to each PE. Replicated input values can be multicast when they are being
  fetched into the buffers.
     
\item {\bf Output halos:} The output buffers at each PE are sized to be slightly
  larger than $K_c\times W_t\times H_t$ to accommodate the halos. The halos
  now contain incomplete partial sums that must be communicated to neighbor PEs for
  accumulation. This communication occurs at the end of computing each output-channel group.
    
\end{itemize}

Our \dflowdense dataflow uses output halos, though the efficiency difference
between the two approaches is minimal.

\begin{figure}[tbp]
\footnotesize
\begin{verbatim}
    BUFFER wt_buf[C][Kc*R*S/F][F];
    BUFFER in_buf[C][Wt*Ht/I][I];
    BUFFER acc_buf[Kc][Wt+R-1][Ht+S-1];
    BUFFER out_buf[K/Kc][Kc*Wt*Ht];
(A) for k' = 0 to K/Kc-1
    {
      for c = 0 to C-1
        for a = 0 to (Wt*Ht/I)-1
        {
(B)       in[0:I-1] = in_buf[c][a][0:I-1];
(C)       for w = 0 to (Kc*R*S/F)-1
          {
(D)        wt[0:F-1] = wt_buf[c][w][0:F-1];
(E)        parallel_for (i=0 to I-1) x (f=0 to F-1)
           {
             k = Kcoord(w,f);
             x = Xcoord(a,i,w,f);
             y = Ycoord(a,i,w,f);
(F)          acc_buf[k][x][y] += in[i]*wt[f];
           }
          }
        }
      out_buf[k'][0:Kc*Wt*Ht-1] =
        acc_buf[0:Kc-1][0:Wt-1][0:Ht-1];
    }
\end{verbatim}
\caption{\dflowdense dataflow.}
\label{fig:ptis-dense}
\end{figure}


\subsection{\dflowsparse Dataflow}

\dflowsparse is a natural extension of \dflowdense that exploits
sparsity in the weights and input activations. The dataflow is
specifically designed to operate on compressed-sparse encodings of the
weights and input activations and to produce a compressed-sparse
encoding of the output activations. At a CNN layer boundary, the
output activations of the previous layer become the input activations
of the next layer.  While prior work has proposed a number of
compressed-sparse
representations~\cite{EIEISCA:2016,CnvlutinISCA:2016,CambriconX:2016},
the specific format used is orthogonal to the sparse architecture
itself. What is key is that decoding a sparse format ultimately yields
a non-zero data value and an index indicating the coordinates of the
value in the weight or input activation matrices.



%
%
%
%

To facilitate easier decoding of the compressed-sparse blocks, weights
are grouped into compressed-sparse blocks at the granularity of an
output-channel group, with of $K_c{\times}R{\times}S$ weights encoded into one
compressed block. Likewise, input activations are encoded at the
granularity of input channels, with a block of $W_t{\times}H_t$
encoded into one compressed block.  At each access, the weight buffer
delivers a vector of $F$ {\bf non-zero} filter weights along with each
of their coordinates within the $K_c{\times}R{\times}S$
region. Similarly, the input buffer delivers a vector of $F$ {\bf
  non-zero} input activations along with each of their coordinates
within the $W_t{\times}H_t$ region. Similar to the dense dataflow, the
multiplier array computes the full cross-product of $F{\times}I$
partial sum outputs, with no extraneous computations.  Unlike a dense
architecture, output coordinates are not derived from loop indices in a
state machine but from the coordinates of non-zero values embedded in the
compressed format.





Even though calculating output coordinates is trivial, the multiplier outputs
are not typically contiguous as they are in \dflowdense. Thus the
$F{\times}I$ multiplier outputs must be scattered to discontiguous
addresses within the $K_c{\times}W_t{\times}H_t$ output range. Because
any value in the output range can be non-zero, the accumulation buffer
must be kept in a dense format.  In fact, output activations will
probabilistically have high density even with a very low density of
weights and input activations, until they pass through a ReLU
operation.

%
%
%
%
%

To accommodate the needs of accumulation of sparse partial sums, we
modify the monolithic $K_c{\times}W_t{\times}H_t$ accumulation buffer
from the \dflowdense dataflow into a distributed array of smaller
accumulation buffers accessed via a scatter network which can be
implemented as a crossbar switch.  The scatter network routes an array
of $F{\times}I$ partial sums to an array of $A$ accumulator banks
based on the output index associated with each partial sum.  Taken
together, the complete accumulator array still maps the same
$K_c{\times}W_t{\times}H_t$ address range, though the address space is
now split across a distributed set of banks. \dflowsparse can be
implemented via small adjustments of
Figure~\ref{fig:ptis-dense}. Instead of a dense vector fetches, (B)
and (D) fetch the compressed sparse input activations and weights,
respectively. In addition, the coordinates of the non-zero values in the
compressed-sparse form of these data structures must be fetched from
their respective buffers (not shown). After that the accumulator buffer
(F) must be indexed with the computed output coordinates from the
sparse weight and activations. Finally, when the computation for the
output-channel group has been completed the accumulator buffer is drained
and compressed into the next level of buffer.




%% file: tex/arch.tex
\section{SCNN Architecture}
\label{sec:arch}

\begin{figure}[tbp]
\centering
\includegraphics[width=\colwidth]{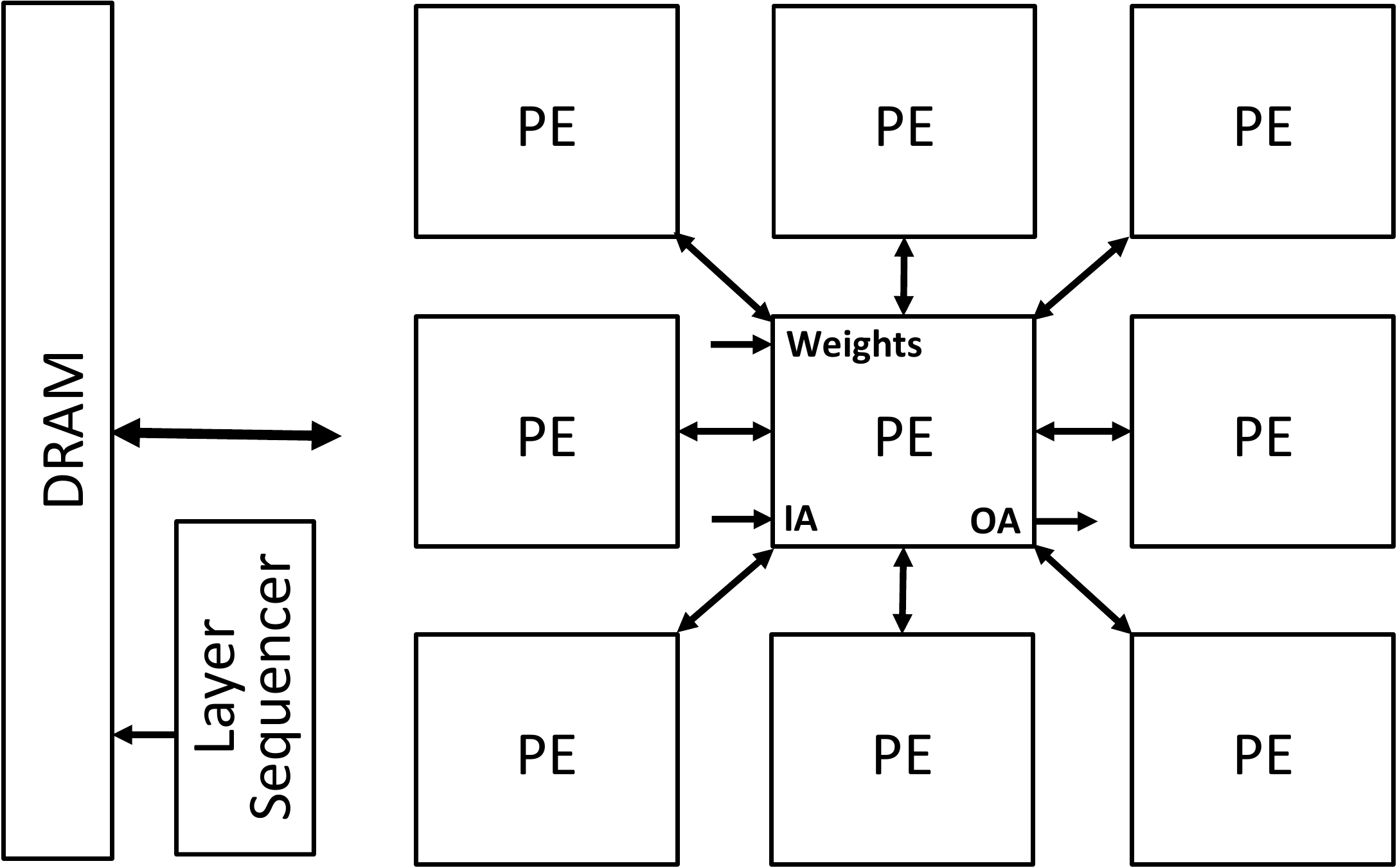}
\caption{Complete SCNN architecture.}
\label{fig:scnn}
\end{figure}

A complete SCNN accelerator employing the \dflowsparse dataflow of
Section~\ref{sec:dataflow} consists of multiple SCNN processing
elements (PEs) connected via simple
interconnections. Figure~\ref{fig:scnn} shows an array of PEs, with
each PE including channels for receiving weights and input
activations, and channels delivering output activations. The PEs are
connected to their nearest neighbors to exchange halo values during
the processing of each CNN layer. The PE array is driven by a layer
sequencer to orchestrate the movement of weights and activations and
is connected to a DRAM controller that can broadcast weights to the
PEs and stream activations to/from the PEs\@.

%
%

\begin{figure*}[tbp]
\centering
\includegraphics[width=0.7\textwidth]{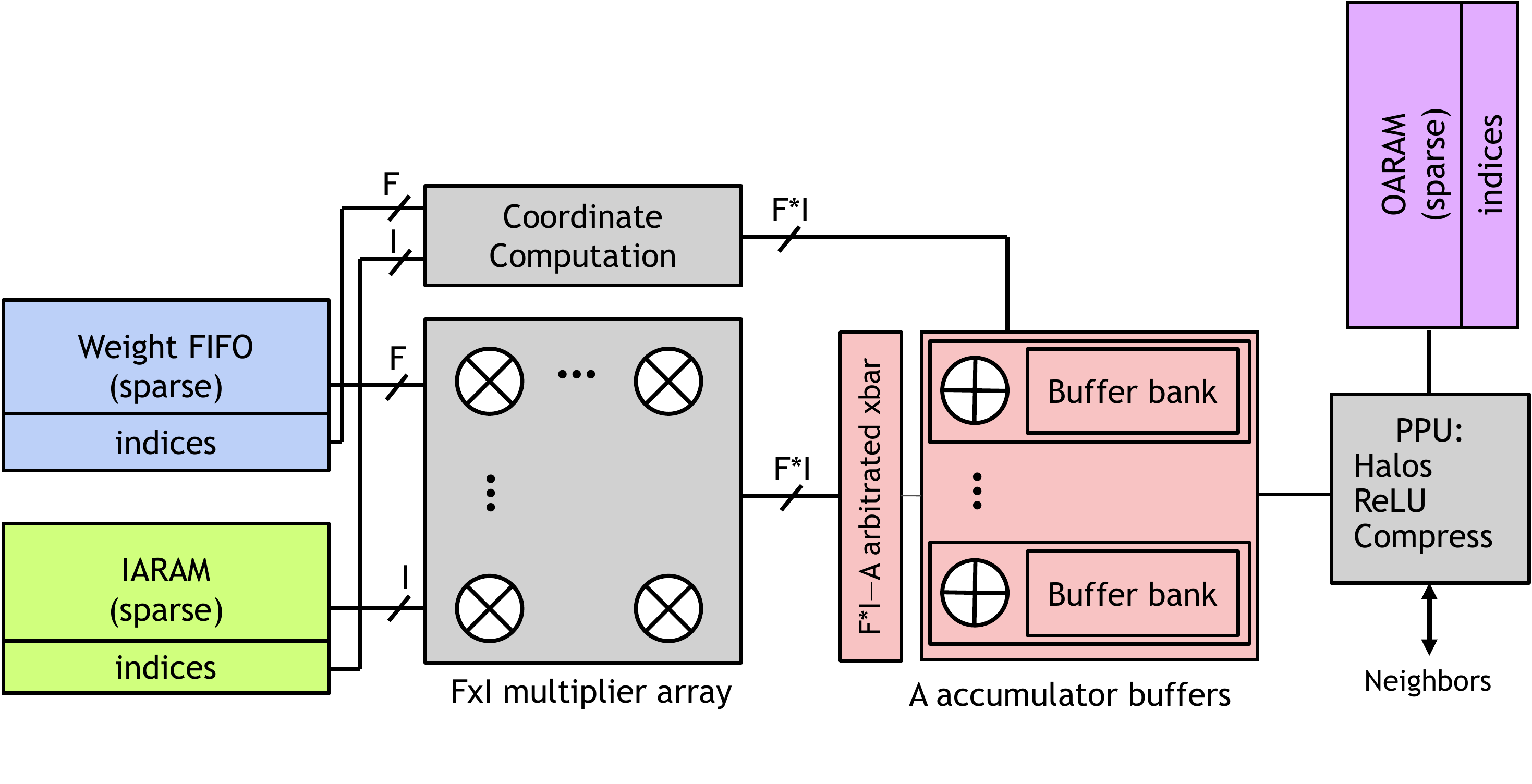}
\caption{SCNN PE employing the \dflowsparse dataflow.}
\label{fig:scnn-pe}
\end{figure*}

{\bf Processing Element (PE) Architecture.}  Figure~\ref{fig:scnn-pe}
shows the microarchitecture of an SCNN PE, including a weight buffer,
input/output activation RAMs (IARAM and OARAM), a multiplier array, a scatter
crossbar, a bank of accumulator buffers, and a post-processing unit
(PPU)\@. To process the first CNN layer, the layer sequencer streams a
portion of the input image into the IARAM of each PE and broadcasts
the compressed-sparse weights into the weight buffer of each
PE\@. Upon completion of the layer, the sparse-compressed output
activation is distributed across the OARAMs of the PEs\@.  When
possible, the activations are held in the IARAMs/OARAMs and are never
swapped out to DRAM\@. If the output activation volume of a layer can
serve as the input activation volume for the next layer, the IARAMs
and OARAMs are logically swapped between the two layers' computation
sequences.

Each PE's state machine operates on the weight and input activations
in the order defined by the \dflowsparse dataflow to produce an
output-channel group of
$K_c\times W_t\times H_t$ partial sums inside the accumulation
buffers.  First, a vector $F$ of compressed weights and a vector $I$
of compressed input activations are fetched from their respective
buffers. These vectors are distributed into the $F{\times}I$
multiplier array which computes a form of the cartesian product of the
vectors.  At the same time, the indices from the sparse-compressed
weights and activations are processed to compute the coordinates in
dense output activation.  The $F{\times}I$ products are delivered to
an array of $A$ accumulator banks, indexed by the output
coordinates. To reduce contention among products that hash to the same
accumulator bank, $A$ is set to be larger than $F{\times}I$\@. Our
results show that $A = 2{\times}F{\times}I$ sufficiently reduces
accumulator bank contention. Each accumulator bank includes adders
and small set of entries for the output channels associated with the
output-channel group being processed.  The accumulation buffers are double-buffered
so that one set of banks can be updated by incoming partial sums while
the second set of banks are drained out by the PPU\@.  When the output-channel group
is complete, the PPU performs the following tasks: (1) exchange
partial sums with neighbor PEs for the halo regions at the boundary of
the PE's output activations, (2) apply the non-linear activation
(e.g. ReLU), pooling, and dropout functions, and (3) compress the
output activations into the compressed-sparse form and write them into
the OARAM\@.

\begin{table}[tpb]
\centering
\caption{SCNN design parameters.}
\begin{footnotesize}
\begin{tabular}{|l|r|}
\hline
\multicolumn{1}{|c|}{PE Parameter} & \multicolumn{1}{|c|}{Value} \\
\hline
Multiplier width & 16 bits \\
Accumulator width & 24 bits \\
IARAM/OARAM (each) & 10KB \\
Weight FIFO & 50 entries (500 B) \\
Multiply array ($F{\times}I$) & 4$\times$4 \\
Accumulator banks & 32 \\
Accumulator bank entries & 32 \\
\hline
\multicolumn{1}{|c|}{SCNN Parameter} & \multicolumn{1}{|c|}{Value} \\
\hline
\# PEs & 64 \\
\# Multipliers & 1024 \\
IARAM + OARAM data & 1MB \\
IARAM + OARAM indices & 0.2MB \\
\hline
\end{tabular}
\end{footnotesize}
\label{tab:parameters}
\end{table}

SCNN uses a simple compressed-sparse encoding approach based on
run-length encoding scheme. The index vector encodes the number of
zeros between each element in the compressed-sparse data
vector. Determining the coordinates in the accumulator buffer for each
multiplier output requires reading the index vectors for $F$ and $I$
and combining them with the coordinates of portion of the output
activation space currently being processed. Four bits per index allows
for up to 15 zeros to appear between any two non-zero
elements. Non-zero elements that are further apart can have a
zero-value placeholder without incurring any noticable degradation in
compression efficiency.

While SCNN will operate most efficiently when the activations fit in
the on-chip activation RAMs, Large networks, such as VGGNet, require
activations to be saved to and restored from DRAM\@. CNN accelerator
architectures such as SCNN can employ a tiling approach that operates
on a 2D subset of the activation space at a time. Our analysis shows
that for both dense and sparse architectures, the DRAM accesses for
one tile can be hidden by pipelining them in tandem with the
computation of another tile. While we do not present the details of
the tiled approach here, we do account for the DRAM access energy
overhead for tiling in Section~\ref{sec:eval}.



%


{\bf SCNN Architecture Configuration.}  While the SCNN architecture
can be scaled across a number of dimensions,
Table~\ref{tab:parameters} lists the key parameters of the SCNN design
we explore in this paper.  The design employs an 8$\times$8 array of
PEs, each with a 4$\times$4 multiplier array, and an accumulator buffer with 32
banks. We chose a design point of 1,024 multipliers to match the
expected computation throughput required to process HD video in
real-time at acceptable frame rates.  The IARAM and OARAM are sized so
that the sparse activations of AlexNet and GoogleNet can fit entirely
within these RAMs so that activations need not spill to DRAM\@. The
weight FIFO and the activation RAMs each carry a 10-bit overhead for
each 16-bit value to encode the coordinates in the compressed-sparse
format.  In total, the SCNN design includes a total of 1,024
multipliers and 1MB of activation RAM\@. At the synthesized clock
speed of the PE of slightly more than 1 GHz, this design achieves a
peak throughput of 2 Tera-ops (16-bit multiplies plus 24-bit adds).





\input{tex/synthesis.tex}

%% file: tex/synthesis.tex



{\bf Area Analysis.}  To prototype the SCNN architecture, we designed
an SCNN PE in synthesizable SystemC and then used the Catapult
high-level synthesis (HLS) tool~\cite{Mentor:Catapult,Martin:HLS:2009}
to generate Verilog RTL.  During this step, we used HLS design
constraints to optimize the design by mapping different memory
structures to synchronous RAMs and latch arrays and pipelining the
design to achieve full throughput. We then used Synopsys Design
Compiler to perform placement-aware logic synthesis and obtain
post-synthesis area estimates in a TSMC 16nm FinFET
technology. Table~\ref{tab:SCNN_area} summarizes the area of the major
structures of the SCNN PE\@. A significant fraction of the PE area is
contributed by memories (IARAM, OARAM, accumulator buffers), which
consume 57\% of the PE area, while the multiplier array only consumes
6\%. IARAM and OARAM are large in size and consume 25\% of the PE
area.  Accumulator buffers, though smaller in size compared to
IARAM/OARAM, are heavily banked (32 banks), contributing to their
large area.

\begin{table}[tpb]
\centering
\caption{SCNN PE area breakdown.}
\begin{footnotesize}
\begin{tabular}{|l|c|c|}
\hline
\multicolumn{1}{|c|}{PE Component} & Size & Area ($mm^2$)\\
\hline
\hline
IARAM + OARAM & 20 KB & 0.031 \\
Weight FIFO & 0.5 KB & 0.004 \\
Multiplier array & 16 ALUs & 0.008 \\
Scatter network & 16$\times$32 crossbar & 0.026 \\
Accumulator buffers & 6 KB & 0.036 \\
Other & --- & 0.019 \\
\hline
Total & --- & 0.123 \\
\hline
\hline
Accelerator total & 64 PEs & 7.9 \\
\hline
\end{tabular}
\end{footnotesize}
\label{tab:SCNN_area}
\end{table}

%% file: tex/infrastructure.tex
\section{Experimental Methodology}

{\bf CNN performance and power measurements.} To model the
performance of the \scnn architecture, we rely primarily on a custom-built
cycle-level simulator. This simulator is parameterizable across dimensions
including number of processing element (PE) tiles, RAM capacity, multiplier
array dimensions ($F$ and $I$), and accumulator buffers ($A$)\@. The \scnn
simulator is driven by the pruned weights and sparse input activation maps
extracted from the Caffe Python interface (\texttt{pycaffe})~\cite{caffe} and
executes each layers of the network one at a time.  As a result, the simulator
captures the effects of the sparsity of the data and its effect on load
balancing within the \scnn architecture.

We also developed TimeLoop, a detailed analytical model for CNN
accelerators to enable an exploration of the design space of dense and
sparse architectures. TimeLoop can model a wide range of data flows,
including \dflowdense, \dflowsparse and others. Architecture
parameters to TimeLoop include the memory hierarchy configuration
(buffer size and location), ALU count and partitioning, and
dense/sparse hardware support. TimeLoop analyzes the input data
parameters, the architecture, and the dataflows, and computes the
number of cycles to process the layer based on a bottleneck analysis
and the counts of ALU operations and accesses to different buffers in
the memory hierarchy. We apply an energy model to the time loop events
derived from the synthesis modeling to compute the overall energy
required to execute the layer. TimeLoop also computes the overall area
of the accelerator based on the inputs from the synthesis
modeling. For SCNN, the area model includes all of the elements from
the synthesizable SystemC implementation. For dense architectures,
area is computed using area of the major structures (RAMs, ALUs, and
interconnect) derived from the SystemC modeling.

{\bf Architecture configurations.}  Table~\ref{tab:configurations}
summarizes the major accelerator configurations that we explore,
including both dense and sparse accelerators. All of the accelerators
employ the same number of multiply ALUs so that we can compare the
performance of the accelerators with the same computational
resources. The dense {\tt DCNN} accelerator operates solely on dense
weights and activations and employs a dataflow called \dflowdcnn,
which is a variant of the \dflowdense data flow described
in Section~\ref{sec:dataflow} that uses a {\em dot-product} as its inner
core operation.  The optimized {\tt DCNN-opt}
architecture has the same configuration as {\tt DCNN} but employs two
optimizations: (1) compression/decompression of activations as they
are transferred out of/into DRAM, and (2) multiply ALU gating to save
energy when a multiplier input is zero. The {\tt DCNN} architecture is
configured with 2MB of SRAM for holding inter-layer activations, and
can hold all of them for AlexNet and GoogLeNet.  The {\tt SCNN}
configuration matches the architecture described in
Section~\ref{sec:arch}, and includes a total of 1MB of IARAM +
OARAM\@. Because the activations are compressed, this capacity enables
all of the activation data for the two networks to be held on chip,
without requiring DRAM transfers for activations.  The larger VGGNet
requires the activation data to be transferred in and out of DRAM\@.
The area required for each accelerator is listed in the last
column. While SCNN has smaller activation RAM capacity, its larger
size is due to the accumulator buffers, as described in
Section~\ref{sec:arch}.

\begin{table}[tpb]
\centering
\caption{CNN accelerator configurations.}
\begin{footnotesize}
\begin{tabular}{|c|c|c|c|c|}
\hline
& \# PEs & \# MULs & SRAM & Area ($mm^2$)\\
\hline
\hline
DCNN & 64 & 1024 & 2MB & 5.9 \\
DCNN-opt & 64 & 1024 & 2MB & 5.9 \\
SCNN & 64 & 1024 & 1MB & 7.9 \\
\hline
\end{tabular}
\end{footnotesize}
\label{tab:configurations}
\end{table}

%

{\bf Benchmarks.}  To explore the sensitivity of the architectures to sparsity
parameters, we employ a synthetic network where we can adjust the degree of
sparsity of both weights and activations.  As described in
Section~\ref{sec:motivation}, we use AlexNet and GoogLeNet for the bulk of our
experiments. With respect to GoogLeNet, we primarily focus on the convolutional
layers that are within the \emph{inception} modules~\cite{googlenet}.  VGGNet
is known to be over-parameterized, which results in an extremely large amount
of inter-layer activation data (6 MB or about 4$\times$) the largest GoogLeNet
layer. However, we use VGGNet as a proxy for large input data (such has
high-resolution images) to explore the implications of tiling data on the SCNN
architecture in Section~\ref{sec:eval:largeNW}.

%

%% file: tex/eval.tex
\section{Evaluation}
\label{sec:eval}


This section first evaluates the sensitivity of SCNN to the sparseness of
weights and activations using a synthetic CNN benchmark. We then discuss the
performance and energy-efficiency of SCNN versus a dense CNN accelerator,
using real world CNN applications. For brevity, all the inception modules in
GoogLeNet are denoted as \texttt{IC\_id} in all of the figures discussed in this
section.


\subsection{Sensitivity to CNN Sparsity}

\begin{figure}[tbp] \centering
\subfloat[Performance]{
\includegraphics[width=\colwidth]{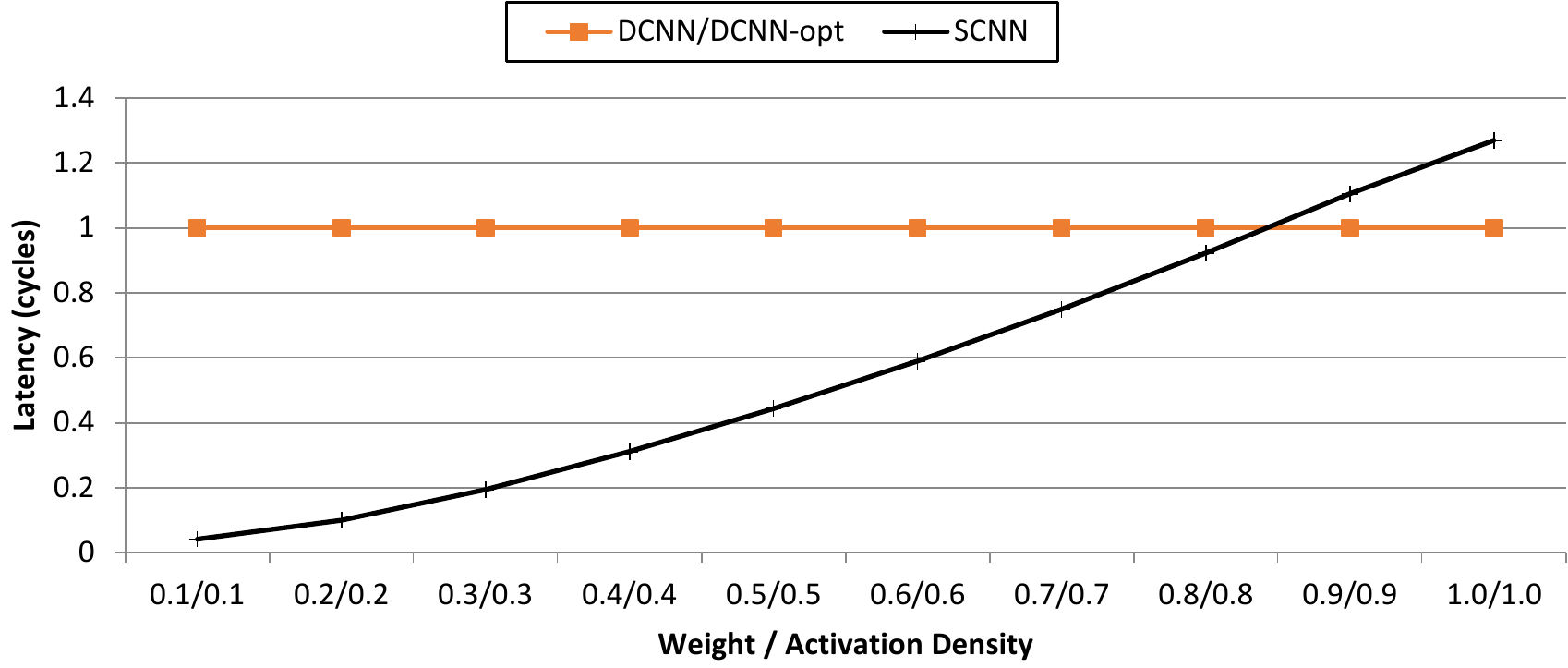}
\label{fig:micro_googlenet_perf}
}
\vspace{0em}
\subfloat[Energy]{
\includegraphics[width=\colwidth]{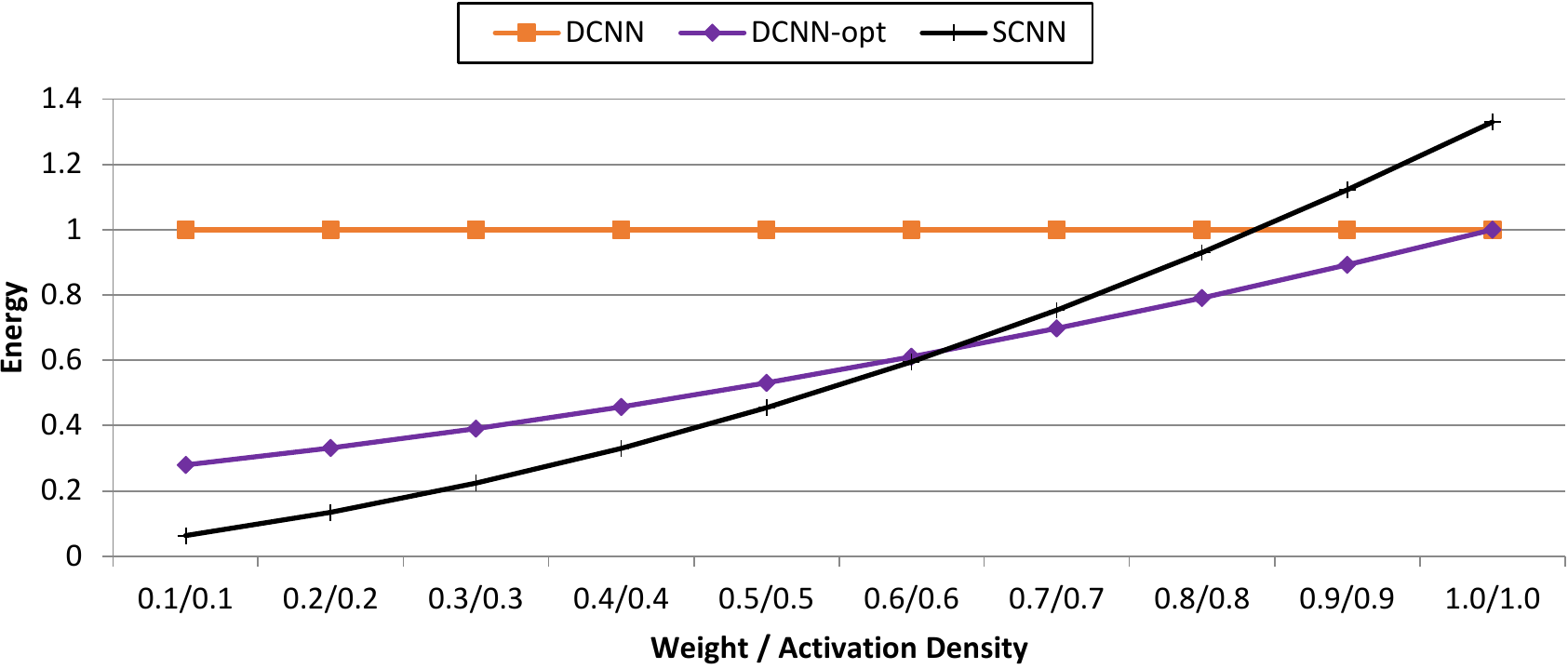}
\label{fig:micro_googlenet_energy}
}
\caption{GoogLeNet performance and energy as a function of density.}
\label{fig:micro} 
\end{figure}

To explore the efficacy of sparse versus dense architecture, we first
measure performance and energy as a function of density of weights and
activations. Using the TimeLoop analysis tool, we examine GoogLeNet as
we artificially sweep the weight and activation densities together
from 100\% (fully dense) down to 10\%. Figure~\ref{fig:micro} shows
results on SCNN, DCNN, and DCNN-opt. On the x-axis, the total density
is the product of the weight and activation density. For example,
0.5/0.5 represents the point where both weights and activations are 50\%
dense so the overall density is 25\%.
Figure~\ref{fig:micro_googlenet_perf} shows that at 100\% density,
SCNN achieves about 79\% of the performance of DCNN because it suffers
from multiplier underutilization (discussed shortly). SCNN starts to
perform better than DCNN as density decreases to 85\%,
reaching a 24$\times$ improvement at 10\% weight/activation density.  This
chart does not include DCNN-opt as the energy optimizations over DCNN
do not affect performance.

Figure~\ref{fig:micro_googlenet_energy} first shows that DCNN-opt's
energy optimizations of zero gating and DRAM traffic compression cause
it to be better than DCNN at every level of density.  These energy
optimizations are surprisingly effective given that they have such a
small effect on the design of the accelerator. At high density, SCNN
is notably less energy efficient than either DCNN architecture due to
the overheads of storing and maintaining the sparse data
structures. SCNN becomes more efficient than DCNN at about 83\% weight/activation
density and more efficient than DCNN-opt at 60\% density. Given the density measurements
of the networks in Figure~\ref{fig:density_sol_speedup}, we expect
SCNN to outperform the dense architectures on nearly all of the
layers of the networks we examined and be largely comparable in energy
efficiency vs. DCNN-opt across these layers.

\subsection{SCNN Performance and Energy}
\label{sect:scnn_perf_energy}

\begin{figure}[tbp] \centering
\subfloat[AlexNet]{
\includegraphics[width=\colwidth]{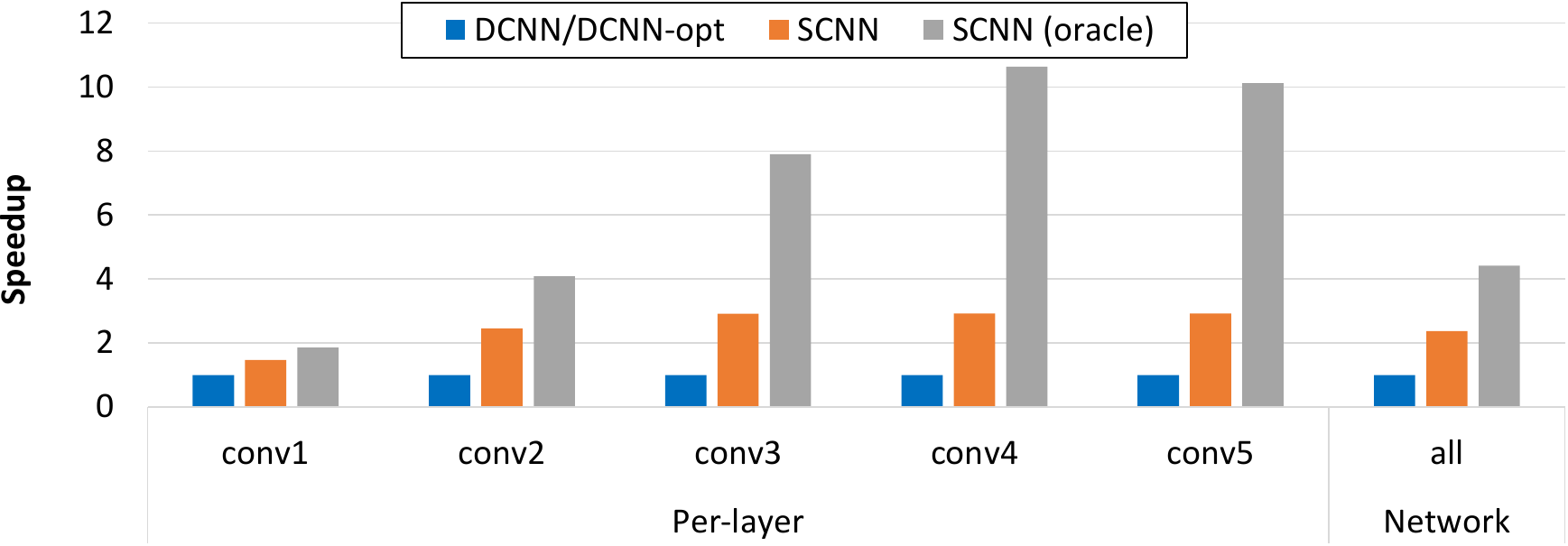}
\label{fig:perf_alexnet}
}
\vspace{0em}
\subfloat[GoogLeNet]{
\includegraphics[width=\colwidth]{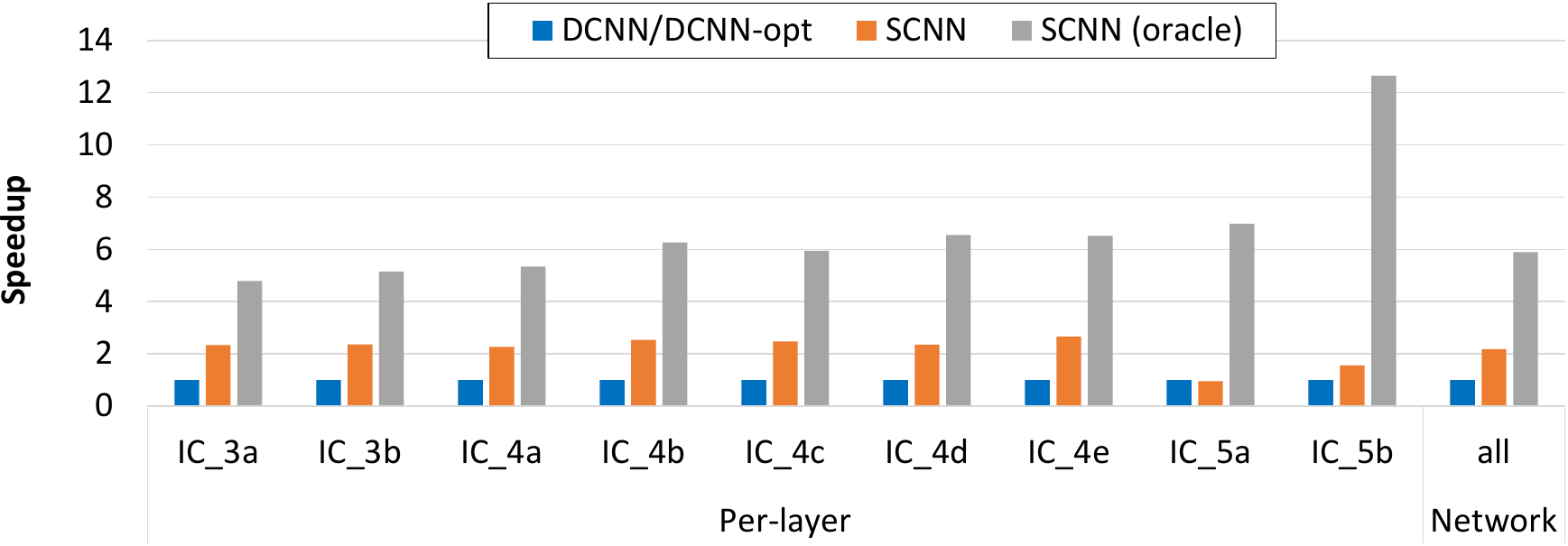}
\label{fig:perf_googlenet}
}
\vspace{0em}
\subfloat[VGGNet]{
\includegraphics[width=\colwidth]{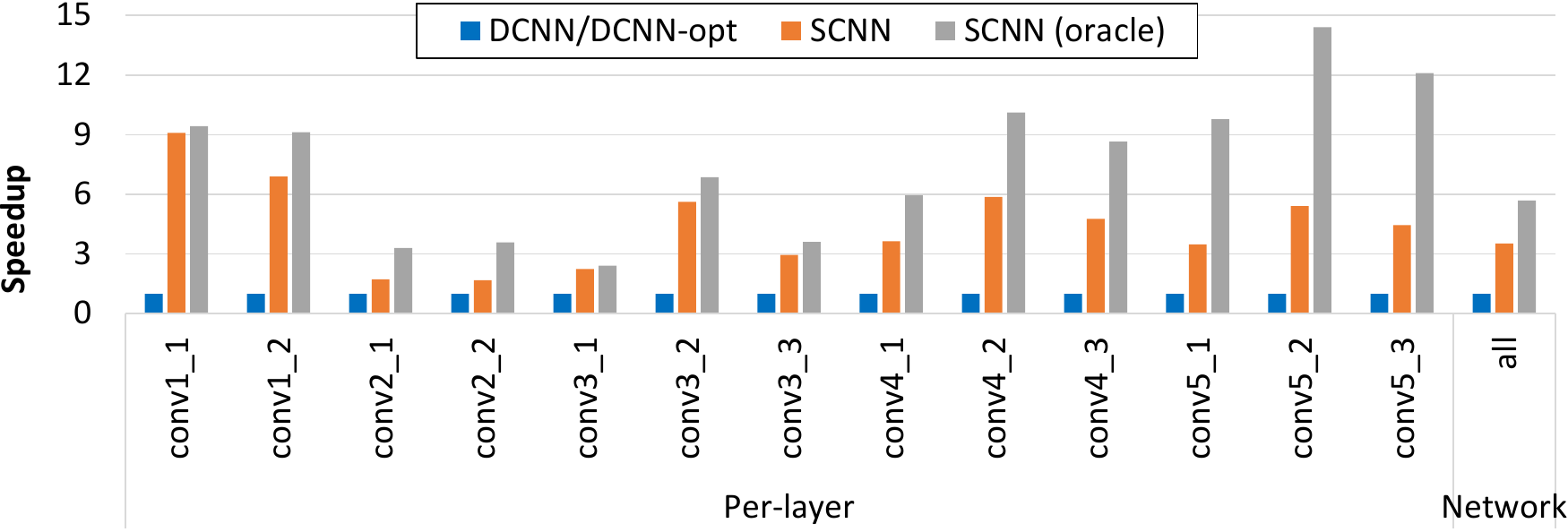}
\label{fig:perf_vgg}
}
\caption{Performance.}
\label{fig:scnn_perf} 
\end{figure}

{\bf Performance.} This subsection discusses the performance
improvements offered by SCNN over the baseline dense DCNN
accelerator. To demonstrate the upper-bound opportunity on
performance, we also present the speedups offered by an
\emph{oracular} SCNN design (\orac). The performance of \orac is
derived by dividing the number of multiplication operations required
for Cartesian product-based convolution (\sect{sec:dataflow}) with the
number of multipliers available on-chip.  \fig{fig:scnn_perf}
summarizes the speedups offered by SCNN versus a dense CNN
accelerator.  Overall, SCNN consistently outperforms the DCNN design
across all the layers of AlexNet, GoogLeNet, and VGGNet, achieving an
average \perfAlexnet, \perfGooglenet, and \perfVggnet network-wide
performance improvement, respectively. 

The performance gap between SCNN versus \orac widens in later layers
of the network, from left to right in the $x$-axis of
\fig{fig:scnn_perf}.  The key reason behind this gap is closely
related to each layer's average PE multiplier array utilization.  In
general, the working set allocated to each of the PEs in the later
layers (e.g., \texttt{IC\_5b}) are smaller than those in the earlier
layers (e.g., \texttt{IC\_3a}).  As a result, it is more difficult to
assign to a PE a sufficient number of non-zero activations and weights
in the later layers to fully utilize the multiplier arrays.

\begin{figure}[tbp] \centering
\subfloat[AlexNet]{
\includegraphics[width=\colwidth]{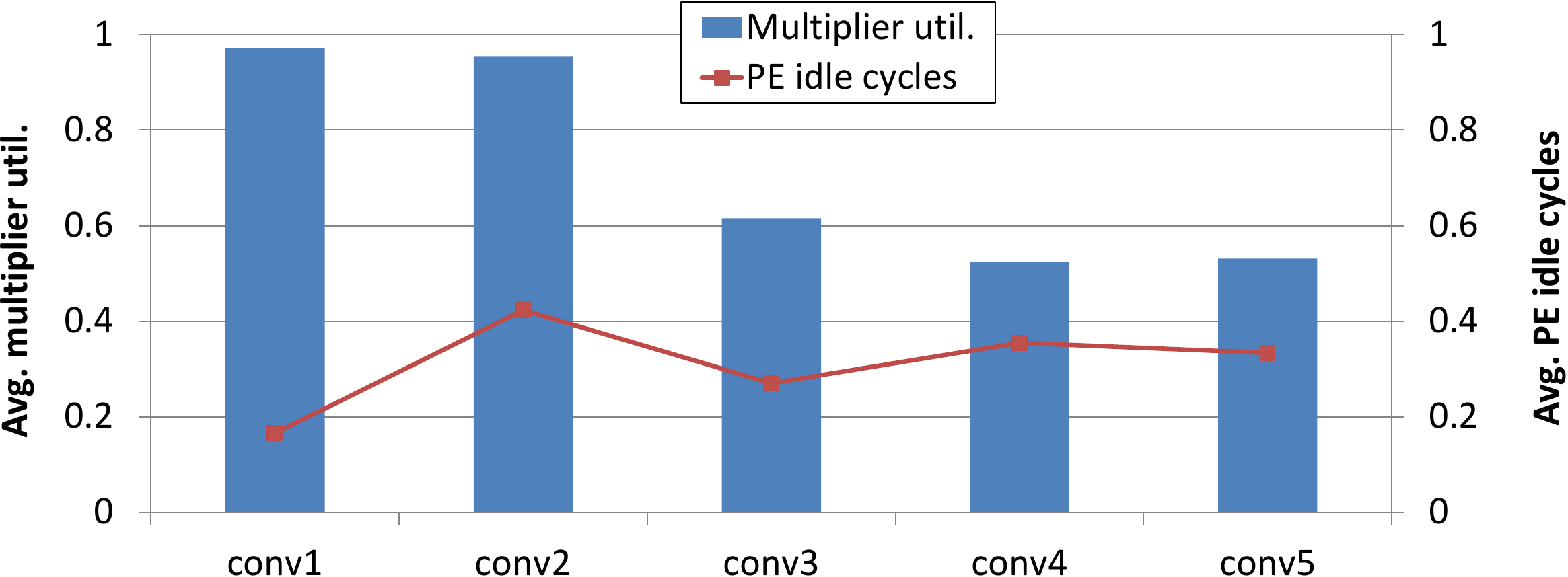}
\label{fig:mul_arr_util_alexnet}
}
\vspace{0em}
\subfloat[GoogLeNet]{
\includegraphics[width=\colwidth]{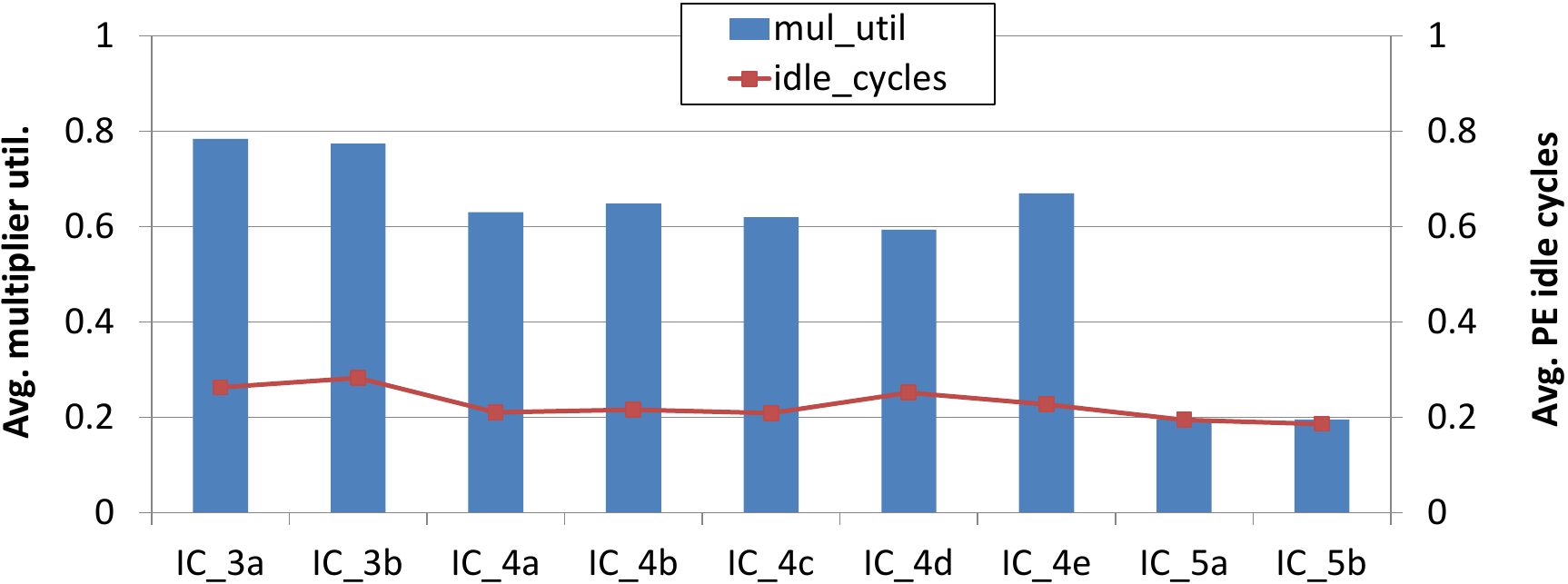}
\label{fig:mul_arr_util_googlenet}
}
\vspace{0em}
\subfloat[VGGNet]{
\includegraphics[width=\colwidth]{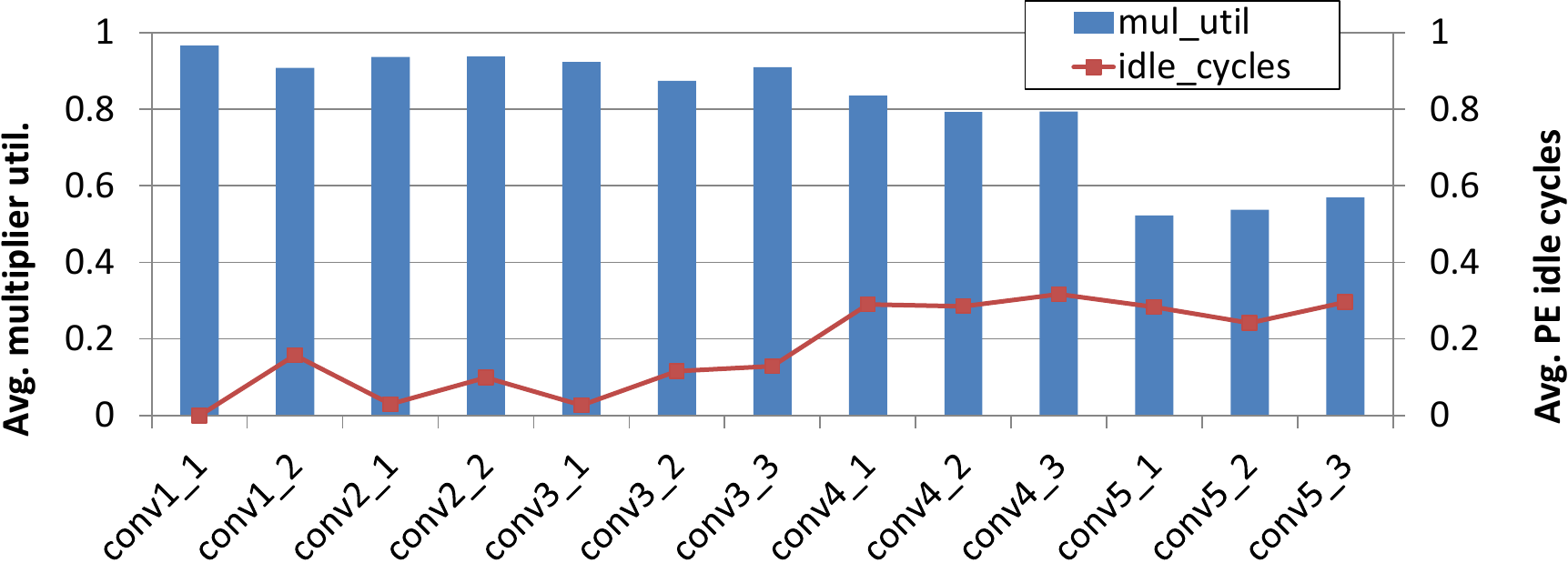}
\label{fig:mul_arr_util_vgg}
}
\caption{Average multiplier array utilization (left-axis) and the average fraction of time PEs are stalled on a global barrier (right-axis), set at the boundaries of output channel groups.}
\label{fig:scnn_utilization} 
\end{figure}

\fig{fig:scnn_utilization} quantitatively demonstrates this intra-PE
fragmentation of the multiplier array.  For the last two inception
modules of GoogLeNet, the fragmentation issue becomes noticeably
severe, with less than an average $20\%$ multiplier utilization. In
this layer $3$ out of the $6$ convolutional layers within the
inception module have a filter size of $1\times1$, only having up to
$8$ non-zero weights within an output-channel group with a $K_c$ value
of $8$.  Nonetheless, the last layers generally account for a small
portion of the overall execution time and SCNN generally provides
significant performance improvement across the network.  The right
y-axis of \fig{fig:scnn_utilization} demonstrates the effect of load
imbalance across the PEs by showing the fraction of cycles spent
waiting at inter-PE barrier.  Overall, although the inter-PE global
barriers and intra-PE fragmentation prevents SCNN from reaching
similar speedups offered by oracular \orac, it still provides an
average \avgPerf performance over DCNN across the three
state-of-the-art CNNs.

\begin{figure}[tbp] \centering
\subfloat[AlexNet]{
\includegraphics[width=\colwidth]{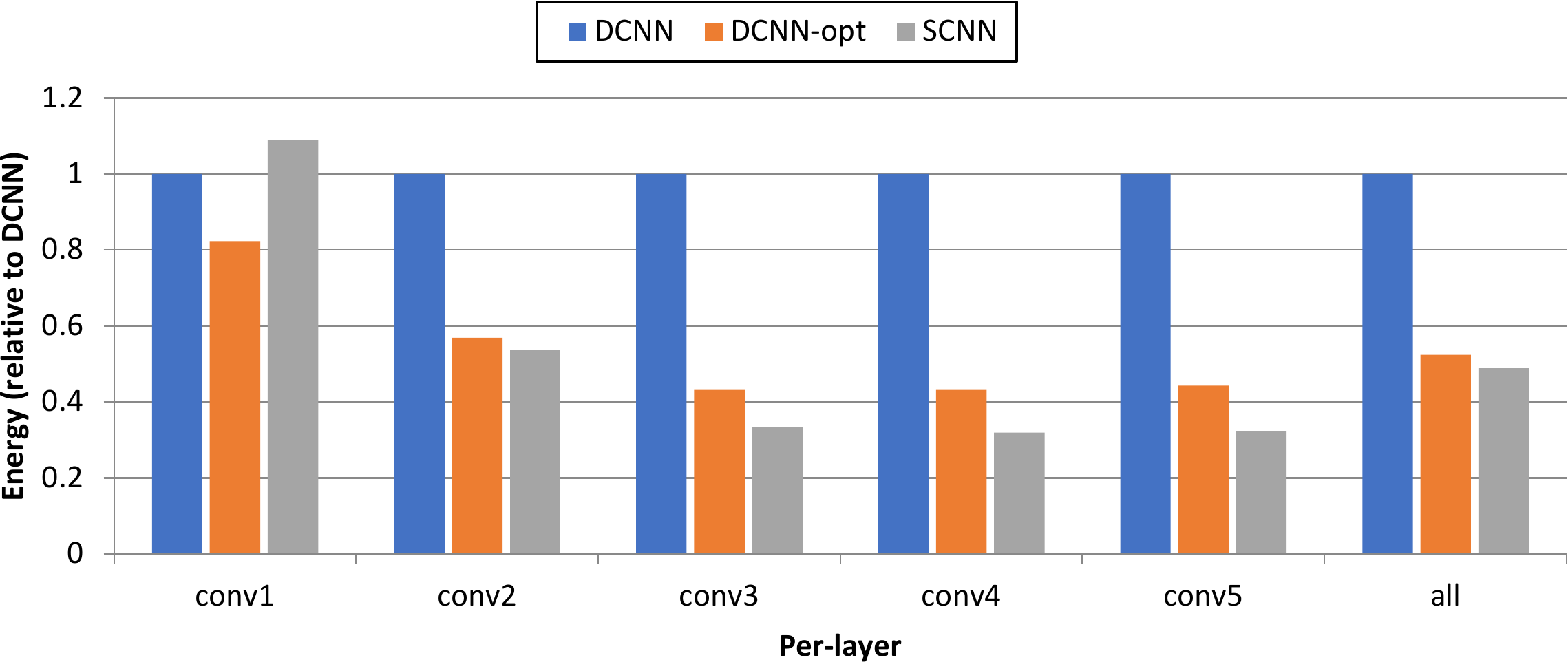}
\label{fig:energy_alexnet}
}
\vspace{0em}
\subfloat[GoogLeNet]{
\includegraphics[width=\colwidth]{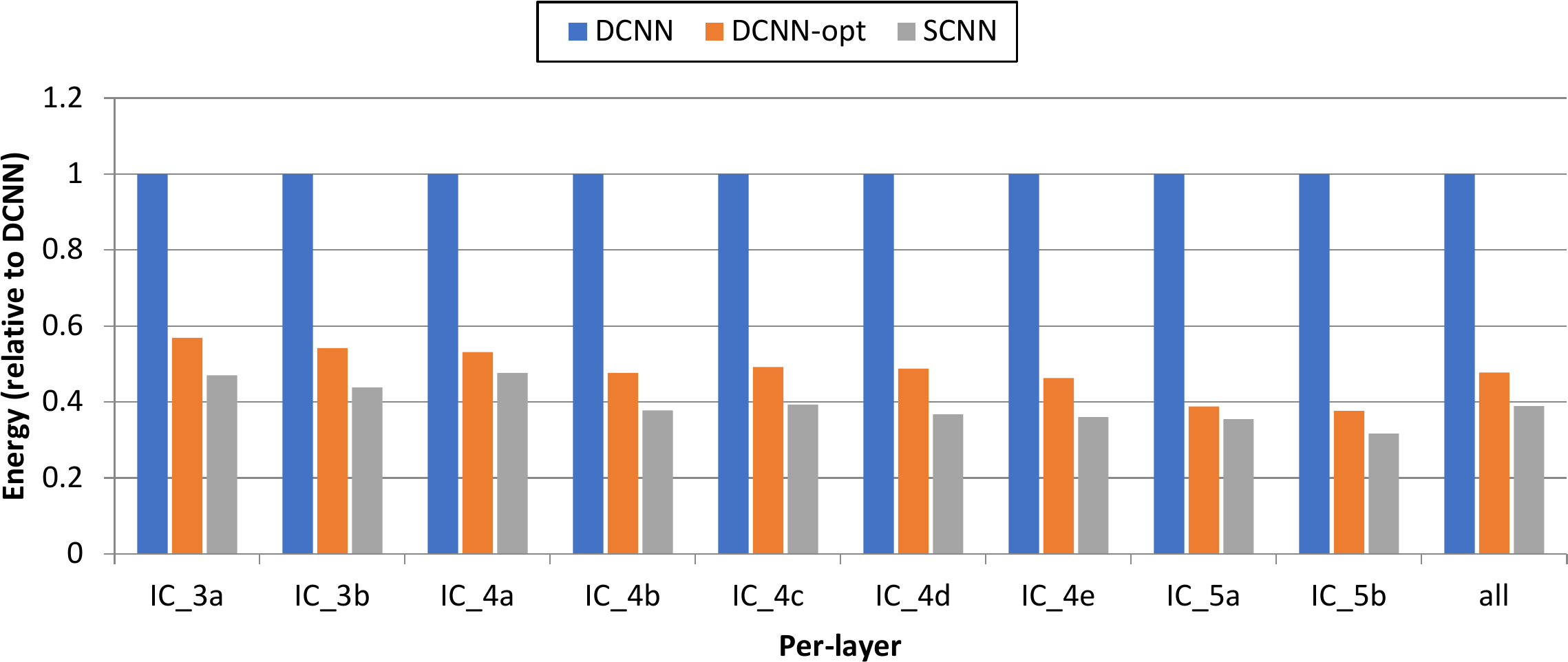}
\label{fig:energy_googlenet}
}
\vspace{0em}
\subfloat[VGGNet]{
\includegraphics[width=\colwidth]{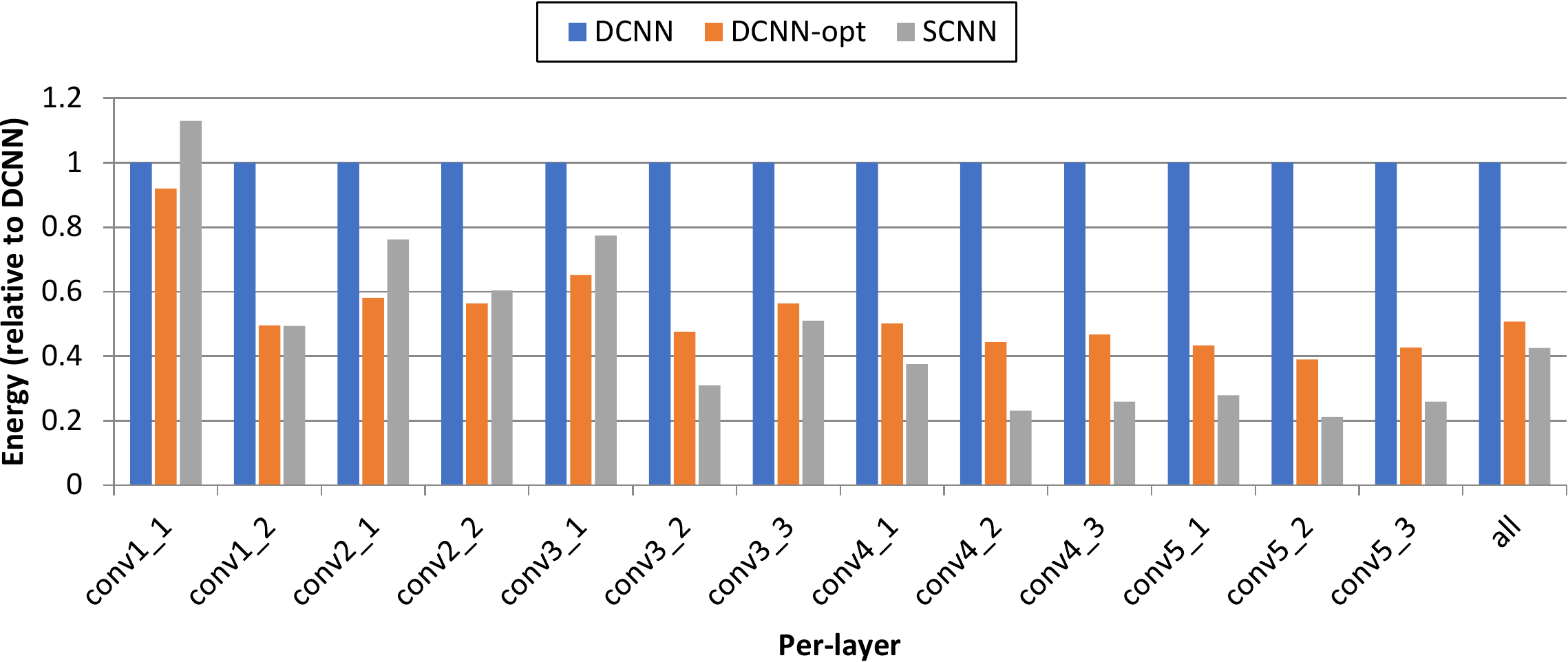}
\label{fig:energy_vgg}
}
\caption{Energy-efficiency.}
\label{fig:scnn_energy} 
\end{figure}

{\bf Energy-efficiency.}
Figure~\ref{fig:scnn_energy} compares the energy of the three
accelerator architectures across the layers of the three
networks. DCNN-opt improves energy efficiency by \avgEnergyDcnnOpt over DCNN,
while SCNN increases average efficiency by \avgEnergyScnn. The
behavior across the layers varies widely depending on the density of
the layer, ranging from 0.89$\times$ to 4.7$\times$ improvement over DCNN
and 0.76$\times$ to 1.9$\times$ improvement over DCNN-opt. Input layers
such as VGGNet\_conv1\_1 and AlexNet\_conv1 usually present a challenge
for sparse architectures because of their 100\% input activation
density. In such cases, the overheads of SCNN's structures such as the
crossbar and distributed accumulation RAMs overshadow any benefits
from fewer arithmetic operations and data movement.

\subsection{PE Granularity}
\label{sect:eval_sensitivity}

As outlined in \sect{sect:scnn_perf_energy}, both cross-PE global
barriers and intra-PE multiplier array fragmentation can contribute to
degradation in the performance of SCNN\@.  We quantify the effects of
both of these factors on system performance by conducting the
following sensitivity study. Assuming a fixed, chip-wide math
throughput of 1,024 FLOPS, we sweep the total number of PEs on-chip
from 64 ($8\times8$ PEs, $16$ multipliers per PE) down to 4
($2\times2$ PEs, $256$ multipliers per PE)\@. Clearly, an SCNN with
$4$ PEs can better sustain the effects of the global barriers than an
SCNN with $64$ PEs\@. However, the $4$ PE configuration is also more
likely to suffer from intra-PE fragmentation because each PE must now
process a larger working set to fully utilize the math units.  When
evaluated on GoogLeNet, SCNN with $64$ PEs achieves an $11\%$ speedup
than the one with $4$ PEs as it does a better job utilizing the math
arrays (average $59\%$ math utilization versus $35\%$). We observed
similar trends for AlexNet and VGGNet, concluding that addressing
intra-PE fragmentation is more critical than inter-PE barriers for
system-wide performance under our \dflowsparse dataflow.




\subsection{Larger Networks}
\label{sec:eval:largeNW}

SCNN is at its most efficient when it can capture all of the
activations in its IARAM and OARAM\@. For a network with large layers
like VGGNet, there are two choices. First, the designer could choose
to provision SCNN with IARAM and OARAM large enough to hold the
largest layer. While this approach would make processing the large
layers more efficient by avoiding off-chip DRAM accesses, smaller
layers would be penalized by incurring larger on-chip access overheads
to RAMs that are larger than necessary. The second choice is to employ
a tiling approach as described in Section~\ref{sec:arch}. Our results
show that this is only required for 9 of the 72 total evaluated layers,
and for these 9 layers the overall energy penalty for shuttling the
activation data to and from DRAM ranges from 5--62\%, with a mean of
18\%. While the tiling approach is
attractive, we expect that there will be low power deployment
scenarios that will motivate the designers of the neural networks to
size them so that they fit in the on-chip SRAM capacity provided by
the accelerator implementation.

%
%

%% file: tex/related.tex
\section{Related Work}

Previous efforts to exploit sparsity in CNN accelerators have focused
on reducing energy or saving time, which will invariably also save
energy.  Eliminating the multiplication when an input operand is zero
is a natural way to save energy.  Eyeriss~\cite{EyerissISSCC:2016}
gates the multiplier when it sees an input activation of zero.  For
non-pruned networks the sparsity of weights is a less significant
factor, and Eyeriss opted not to gate the multiplier on zero weights.
This approach will save energy, but not save cycles.  SCNN also saves
energy by eliminating all the unnecessary multiplications, but doesn't
even prepare to do a multiplication when any input operand is zero,
thus saving time as well.

Another approach to reducing energy is to reduce data transfer costs
when the data is sparse.  Eyeriss uses a run length encoding scheme
when transferring activations to and from DRAM.  This saves energy
(and time) by reducing the number of DRAM accesses.  However, after
the data is loaded into the on-chip buffer the data is stored in
expanded form.  Thus, there is no savings on data transfers from one
internal buffer to another internal buffer or to the multipliers.
SCNN also uses a compressed representation for all data coming from
DRAM, but also maintains that compressed representation in the on-die
buffers.

Another CNN accelerator designed to exploit sparsity,
Cnvlutin~\cite{CnvlutinISCA:2016} compresses activation values.
However, it does not compress weights.
Cambricon-X~\cite{CambriconX:2016}, does not compress the data coming
from DRAM.  
Nor does it keep activations in a compressed form in the internal buffers, 
except in the queues directly delivering activations to the multipliers.  
Cambricon-X does keep only non-zero weights in the internal buffers.
In contrast, 
SCNN keeps both weights and activations in a compressed form in both DRAM and internal buffers.
This saves data transfer time and energy on all data transfers and
allows the chip hold larger models for a given amount of internal
storage.

Not delivering zero activation values or zero weights to the
multipliers can save time by avoiding cycles when the multiplier has
nothing to do.  Cnvlutin selects only non-zero activation values for
delivery as multiplier operands, but does occupy a multiplier with
zero weights.  Cambricon-X does not deliver either zero activations or
weights to the multipliers.  SCNN also does not deliver either zero
activations or weights to the multipliers.

Finally, the EIE CNN accelerator~\cite{EIEISCA:2016} uses a compressed
representation of both activations and weights, and only delivers
non-zero operands to the multipliers.  However, EIE is designed for
the fully connected layers of a CNN model, while SCNN is targeting the
convolutional layers where the majority of the computations are~\cite{Cong:CNNComputation:2014}.

%% file: tex/conclusion.tex
\section{Conclusion}
\label{sec:conc}

This paper presents the Sparse CNN (SCNN) accelerator architecture for
inference in convolutional neural networks. SCNN exploits sparsity in
both weights and activations using the \dflowsparselong (\dflowsparse)
dataflow. This approach enables SCNN to
use a novel Cartesian product-based computation architecture that
maximizes reuse of weights and activations within a set of distributed
processing elements. In addition, it allows the use of a dense
compressed representation for both weights and activations to be used
through almost the entire processing flow.  This approach reduces data
movement and on-die storage capacity relative to alternative
architectures.  Our results show that the SCNN architecture starts to
beat a dense architecture in performance and energy efficiency when the
weights and activations are each less than 85\% dense. On
three contemporary networks (AlexNet, GoogLeNet, and VGGNet) SCCN
achieves performance improvements over the dense architecture by a
factor of \avgPerf while still being energy-efficient by a factor of
\avgEnergyScnn.